\documentclass[10pt,twocolumn,letterpaper]{article}
\usepackage[pagenumbers]{cvpr}
\usepackage[T1]{fontenc}
\usepackage{mathtools}
\usepackage{multirow}
\usepackage{array}
\usepackage{colortbl}
\usepackage{enumitem}
\definecolor{cvprblue}{rgb}{0.21,0.49,0.74}
\newcommand{\method}{OmniPersist\xspace}

\newcommand{\up}{\ensuremath{\uparrow}}
\newcommand{\dn}{\ensuremath{\downarrow}}

\usepackage[pagebackref,breaklinks,colorlinks,allcolors=cvprblue]{hyperref}

\usepackage{overpic} 
\usepackage{multirow}
\usepackage{makecell}

\newcommand{\firstcolor}{red!30}
\newcommand{\firstcell}[1]{\cellcolor{\firstcolor}#1}
\newcommand{\secondcolor}{orange!30}
\newcommand{\secondcell}[1]{\cellcolor{\secondcolor}#1}
\newcommand{\thirdcolor}{yellow!30}
\newcommand{\thirdcell}[1]{\cellcolor{\thirdcolor}#1}

\title{OmniPersist: Persistent Dynamic World Modeling\\via Coupled Panoramic Evolution}
\author{
Zhilu Zhang\textsuperscript{1}\quad
Xiaoyu Liu\textsuperscript{2}\quad
Wenxi Liu\textsuperscript{3}\quad
Shengfeng He\textsuperscript{1}\thanks{Corresponding author.}\\
\textsuperscript{1}Singapore Management University\quad 
\textsuperscript{2}Harbin Institute of Technology\quad
\textsuperscript{3}Fuzhou University\\[0.5em]
{Project: \href{https://omnipersist.github.io/}{https://omnipersist.github.io/}}
}

\DeclareCaptionFont{promptoverview}{\fontsize{8}{10.5}\selectfont}

\begin{document}

\maketitle
\begin{abstract}

Action-conditioned world models play a vital role in robotics and autonomous driving, yet existing video diffusion models struggle to maintain dynamic object persistence. When the camera turns away and returns, out-of-view moving entities fail to evolve plausibly in latent space, often mutating or vanishing entirely due to bounded perspective views. To resolve this issue, we propose OmniPersist, a unified framework that couples perspective video generation with panoramic evolution within a single-stage diffusion model, without relying on auxiliary models or initial panoramic inputs. To maintain efficiency, we model the panoramic stream at a lower frame rate,  bridging both views across space and time via Geometric Anchoring (GeoAnch) for spatial correspondence and Temporal Compensation (TemComp) for temporal aggregation. OmniPersist is agnostic to camera-control architectures, supporting both latent-level and attention-level conditioning. Extensive experiments on a newly constructed dynamic benchmark across image-to-video and video continuation tasks demonstrate consistent gains in visual quality, camera controllability, revisit consistency, and dynamic persistence.

\end{abstract}

\section{Introduction}
\label{sec:intro}

World models~\cite{team2026dreamx, wang2026matrix, zhu2026sana, 
team2026advancing, gao2026infinite} aim to predict future states from current observations conditioned on actions, 
playing an increasingly important role in robotics~\cite{li2026causal,yang2025physworld} 
and autonomous driving~\cite{li2026drivevla,wang2024drivedreamer}. 
A prevalent paradigm generates future videos via diffusion models~\cite{blattmann2023align,peebles2023scalable,yang2025cogvideox,wan2025wan}, 
guided by textual prompts alongside camera trajectories or discrete motion signals, 
as demonstrated by LingBot-World~\cite{team2026advancing,gao2026infinite}
and Matrix-Game 3.0~\cite{wang2026matrix}. Despite demonstrating compelling visual fidelity through large-scale training,  
these models fundamentally struggle to maintain object persistence in dynamic scenes. 
When the viewpoint shifts away and subsequently returns, 
out-of-view objects fail to evolve plausibly in the latent space. 
As a result, their visual consistency and motion continuity break down, 
even causing returning entities to mutate or vanish entirely.

\begin{figure}[t!]
\centering
\includegraphics[width=\linewidth]{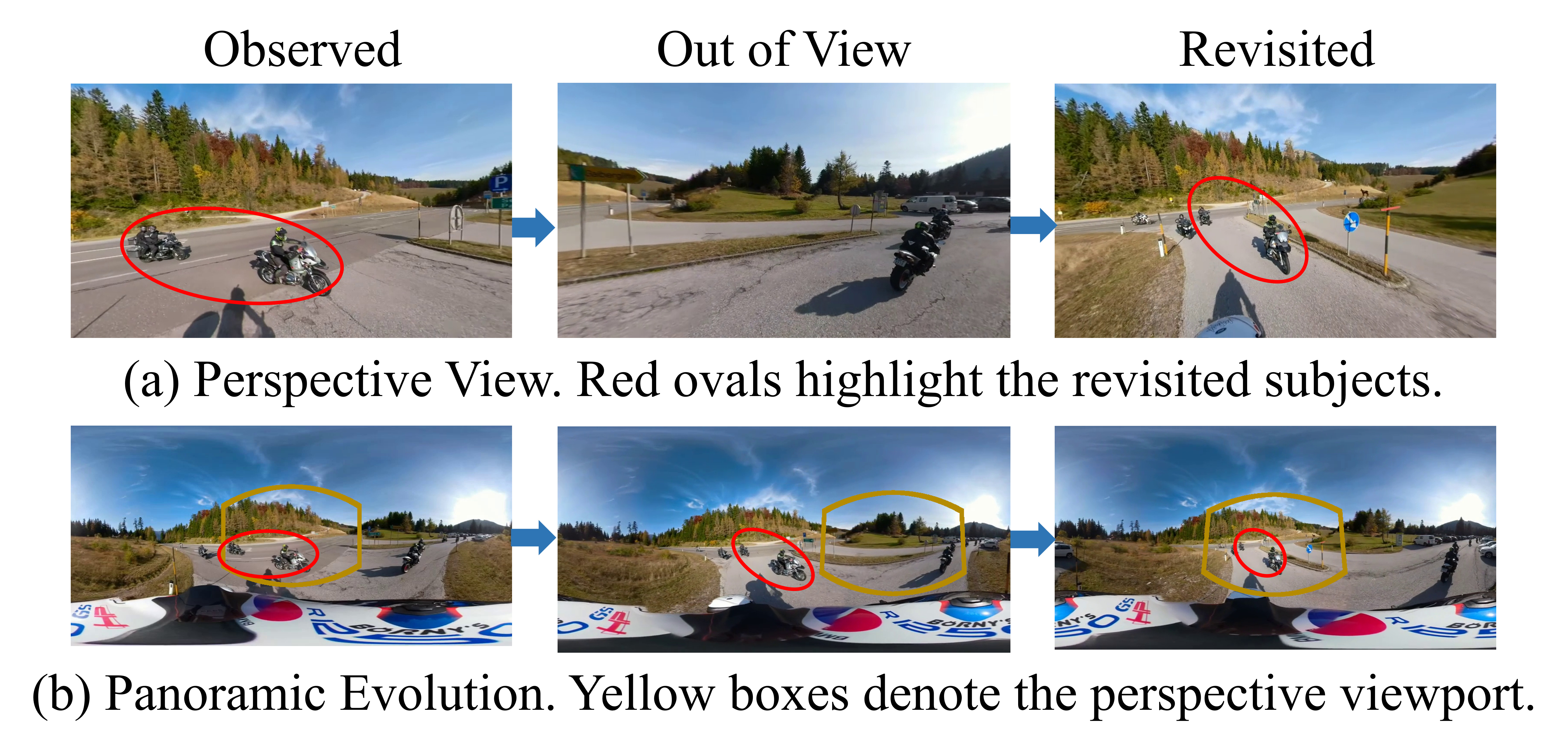}
\vspace{-8mm}
\caption{\textbf{An example of dynamic subjects across a camera revisit.} 
While moving subjects temporarily exit the perspective view, the $360^\circ$ panoramic view continuously captures their motion outside the perspective viewport. 
Best viewed when zoomed in.}
\label{fig:teaser}
\vspace{-4mm}
\end{figure}

We attribute this limitation to both training data biases and fundamental modeling bottlenecks. 
On the data side, current training data for dynamic scenes heavily favor continuous tracking, 
rarely featuring scenarios where moving entities exit and subsequently re-enter 
the camera's field of view (FoV). 
On the modeling side, vanilla diffusion models are inherently limited in 
preserving out-of-view dynamics without dedicated constraints and priors. 
Recent works have recognized this challenge and 
sought to mitigate it mainly from two perspectives. Some methods utilize 
retrieval frames~\cite{chen2026hybrid} 
and compressed implicit contexts~\cite{yu2026memlearner} as memory. 
Others attempt to predict dynamics of specific entity through 
auxiliary generative models~\cite{duan2026liveworld} 
and LLMs~\cite{wang2026worlddirector}. 
However, the former often recall static appearance cues that fail to reflect 
continuous natural motion, while the latter introduce 
heavy system complexity and scale poorly to multi-entity scenarios. 
Furthermore, these methods are designed primarily for object
consistency across temporal chunks. 
In fact, even within a single video chunk, combining intra-window bidirectional attention 
with advanced camera control approaches (\eg, UCPE~\cite{zhang2026ucpe}) remains insufficient 
to sustain out-of-view dynamic evolution.

To address this problem, we propose modeling scene evolution 
in panoramic space, leveraging complete $360^\circ$ spatial context 
to improve the persistence of dynamic world models. 
This paradigm offers two distinct advantages. 
First, conceptually, the omnidirectional nature of panoramic space 
ensures that when an object leaves the camera view, its continuous motion can still be explicitly predicted on a global canvas. 
Moreover, this formulation naturally supports multiple moving objects and inherently preserves background memory, as shown in Fig.~\ref{fig:teaser}. 
Second, practically, panoramic scenes can be represented as standard 2D frames via equirectangular projection (ERP), allowing existing video diffusion foundation models to adapt seamlessly to panoramic data while fully capitalizing on their strong pretrained priors.

Specifically, we propose \textbf{OmniPersist}, which jointly models 
the perspective camera view and its panoramic counterpart within a single video diffusion model. 
This unified framework achieves persistent dynamic world modeling in an end-to-end manner, 
without relying on auxiliary models or additional input. 
Since generating a dense panorama for every perspective frame introduces heavy computational overhead, 
we propose to design the panoramic stream to operate at a lower frame rate. 
To effectively bridge these two representations across space and time, 
we introduce two complementary modules, \ie, \emph{Geometric Anchoring (GeoAnch)} and \emph{Temporal Compensation (TemComp)}. 
In particular, \emph{GeoAnch} establishes explicit geometric correspondences between 
the perspective and panoramic views, enabling direct feature exchange at synchronized timestamps. 
Meanwhile, \emph{TemComp} aggregates features across neighboring frames, 
effectively compensating for the temporal discrepancy caused 
by the differing frame rates. 
Concurrently, AlayaVista~\cite{tan2026alayavista} (within the \textit{past two months}) also explores panoramic representations, but relies on a multi-stage pipeline of panoramic initialization, state evolution, and perspective refinement.
In contrast, OmniPersist avoids generating redundant full-rate panoramas, establishing a single-stage, unified framework for persistent dynamic world modeling.

To ensure broad applicability, OmniPersist is designed to be agnostic 
to underlying camera-control architectures. 
We integrate our framework with two representative paradigms, \ie, 
latent-level trajectory injection via CameraCtrl~\cite{he2025cameractrl} 
and attention-level pose conditioning via UCPE~\cite{zhang2026ucpe}. 
Furthermore, to benchmark dynamic persistence under complex camera movements, 
we construct a paired panoramic--perspective dataset featuring diverse trajectory patterns. 
We systematically evaluate OmniPersist under two settings, 
namely image-to-video (I2V) generation and video continuation. 
Extensive experiments across visual quality, camera controllability, 
revisit consistency, and dynamic persistence demonstrate that 
our coupled panoramic evolution yields consistent gains across both control mechanisms.

In summary, our main contributions are as follows:
\begin{itemize}
    \item We propose \textbf{OmniPersist}, a single-stage unified framework that jointly generates perspective video and temporally sparse panoramic evolution to improve out-of-view dynamic persistence.
    
    \item We introduce \emph{GeoAnch} and \emph{TemComp} to couple dense perspective and sparse panoramic streams through synchronized geometric exchange and neighboring-time feature aggregation.
    
    \item We construct a paired panoramic--perspective dynamic dataset and demonstrate 
    consistent gains in dynamic persistence and revisit consistency across diverse camera-control mechanisms.
\end{itemize}
\section{Related Work}
\label{sec:related}

\noindent\textbf{Camera-Controlled Video Generation.}
Camera trajectory control has become an essential capability for interactive world 
models~\cite{team2026dreamx,wang2026matrix,zhu2026sana,team2026advancing,gao2026infinite}. 
Existing efforts encompass text-to-video~\cite{wang2024motionctrl,bahmani2025vd3d}, 
image-to-video~\cite{li2025realcam,ren2025gen3c,zheng2026versecrafter}, 
and video re-shooting~\cite{bai2025recammaster,yu2025trajectorycrafter,team2026inspatio}, 
and roughly divide into two paradigms.
(1) \emph{Camera-parameter conditioning} directly encodes viewpoints as extrinsic matrices~\cite{wang2024motionctrl}, 
Pl\"ucker rays~\cite{he2025cameractrl,bahmani2025vd3d}, or relative positional embeddings~\cite{li2026rerope,zhang2026ucpe,yin2026scope}. 
These representations are used to modulate feature latents via auxiliary branches~\cite{he2025cameractrl,bahmani2025vd3d} 
or inject geometric biases into attention mechanisms~\cite{zhang2026ucpe,yin2026scope,li2026rerope}. 
These methods operate strictly within the visible frustum, 
offering no spatial substrate to sustain the evolution of out-of-view dynamic entities.
(2) \emph{Explicit 3D reprojection} grounds camera trajectories by warping reconstructed 3D geometry 
into target image or latent spaces. 
For instance, ViewCrafter~\cite{viewcrafter} and GEN3C~\cite{ren2025gen3c} condition diffusion on 
rendered point clouds with iterative scene expansion or 3D caching. 
MosaicMem~\cite{yu2026mosaicmem} retrieves 3D-lifted patches from spatial memory, whereas LSM-World~\cite{wang2026latent} caches 3D latent tokens directly to bypass RGB rendering and re-encoding.
However, these approaches rely heavily on external depth estimators 
and are predominantly restricted to static scenes.
In contrast, \textbf{OmniPersist} bypasses 3D reconstruction and 
provides an omnidirectional canvas that naturally sustains unseen dynamic evolution. 
It acts as an orthogonal, plug-and-play spatial substrate, 
seamlessly integrating with the camera-parameter conditioning paradigm.

\noindent\textbf{Persistent Dynamic World Modeling.}
World models simulate future states conditioned on current 
observations and actions. 
A prevalent paradigm generates future videos guided by text 
and camera trajectories~\cite{wang2026matrix,zhu2026sana,team2026advancing,gao2026infinite}. 
However, maintaining the evolving states of dynamic out-of-view entities remains an unresolved challenge. 
Existing approaches broadly follow three directions.
(1) \emph{Memory-based modeling}. 
HyDRA~\cite{chen2026hybrid} and MemLearner~\cite{yu2026memlearner} query or compress historical 
observations to preserve generation consistency. 
ReMind~\cite{xu2026teaching} employs a node-structured curriculum that compels the model to retrieve relevant historical states. 
While historical memory provides informative anchors, these methods inherently infer unobserved evolution 
from past evidence, often struggling to sustain continuous visual transitions.
(2) \emph{Auxiliary entity tracking}. 
LiveWorld~\cite{duan2026liveworld} uses multiple generative monitors to advance unseen entity states within a persistent scene, 
while WorldDirector~\cite{wang2026worlddirector} employs an LLM to predict object trajectories prior to visual generation. 
However, relying on external entity-level monitors or LLM planners introduces heavy system overhead, 
causes pipeline brittleness, and scales poorly to dense, multi-entity interactive scenarios.
(3) \emph{Panoramic world-state modeling}. 
In concurrent work, AlayaVista~\cite{tan2026alayavista} first evolves panoramic latent states and then synthesizes perspective observations through viewport rendering and refinement stages. 
However, it relies on a separate panorama initialization stage and incurs high computational costs by predicting full-rate panoramas. 
In contrast, \textbf{OmniPersist} derives its panoramic conditions directly from the observed perspective prefix and predicts panoramic states at a lower temporal rate, establishing a single-stage joint diffusion framework.

\noindent\textbf{Panoramic Video Generation.}
Panoramic video generation aims to synthesize temporally coherent $360^\circ$ content, 
typically achieved by adapting pretrained perspective video diffusion models. 
Therein, perspective-to-panorama methods expand input videos while preserving their
content and dynamics~\cite{luo2025beyond,li2026cubecomposer,wu2026360anything,xie2025videopanda}, 
whereas text- and image-conditioned approaches synthesize omnidirectional scenes and motion~\cite{wang2024dvd,zhang2025panodit,fang2026panoramic,xia2026panowan,xie2025videopanda}. 
Recent methods further extend panoramic synthesis to camera-controlled exploration~\cite{li2026panoworld,tan2026mugen} 
and 3D world construction~\cite{wang2025evoworld,yang2025matrix,chen2026pantheon360}. 
Technically, these methods address spherical distortion and boundary seams through ERP-aware designs~\cite{xia2026panowan,wu2026360anything,tan2026mugen}, 
cross-view coherence via multi-view attention~\cite{fang2026panoramic,xie2025videopanda}, 
and exploration consistency through geometric guidance and spatial memory~\cite{wang2025evoworld,yang2025matrix,li2026panoworld,chen2026pantheon360}. 
Collectively, these advances establish the feasibility of using video diffusion models to evolve panoramic states. 
Unlike prior works targeting standalone panoramic synthesis, 
\textbf{OmniPersist} couples an efficient, low-rate panoramic stream 
with perspective rollouts to anchor out-of-view dynamic persistence.

\section{Method}
\label{sec:method}
\subsection{Overview}
\label{sec:overview}

We study persistent dynamic world modeling where dynamic subjects may leave and subsequently re-enter the camera field of view. Given a textual prompt, an observed perspective prefix $\mathbf{X}_{0:K-1}$, and a $T$-frame camera trajectory $\mathbf{C}^{\mathrm{x}}$ with known intrinsics, the objective is to synthesize the remaining frames of the $T$-frame perspective video $\mathbf{X}$. Crucially, returning subjects must maintain identity consistency and plausible physical motion throughout the interval spent outside the viewport.

\begin{figure*}[t]
    \centering
    \includegraphics[width=\textwidth]{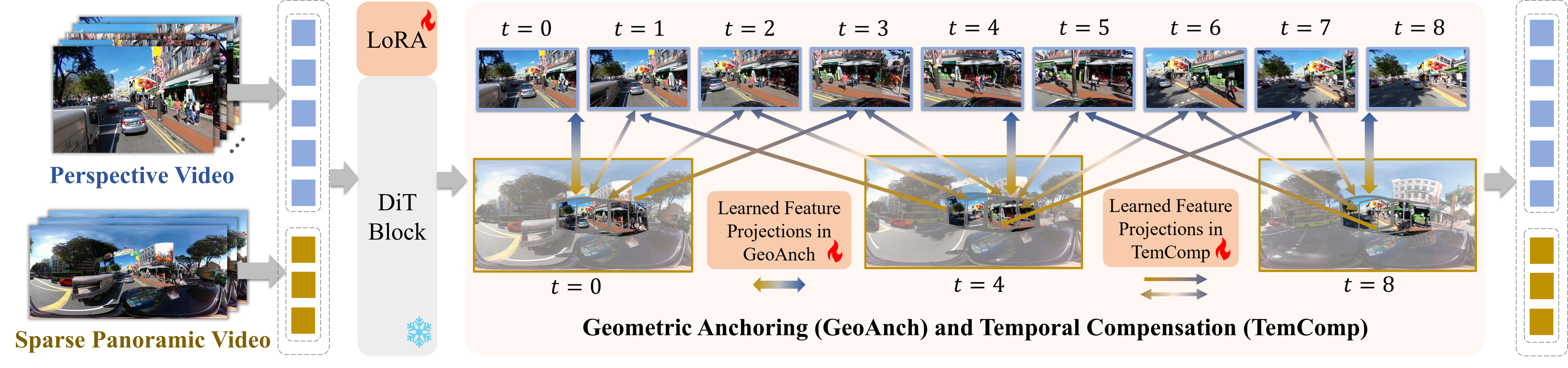}
    \vspace{-6mm}
    \caption{\textbf{Overview of the OmniPersist framework.} 
    Tokens from the dense perspective stream and the temporally sparse panoramic stream 
    are jointly processed by a shared, frozen Diffusion Transformer (DiT) block adapted via LoRA. 
    The two streams coordinate across space and time via dedicated learned feature projections 
    to exchange complementary context. 
    GeoAnch transfers spatially aligned features at synchronized timestamps, while TemComp aggregates features across neighboring timestamps for intermediate observations. 
    For illustration, we depict a temporal sampling interval of $s=4$ in image-space indices.
    }
    \label{fig:framework}
    \vspace{-1mm}
\end{figure*}

OmniPersist addresses this task by jointly generating $\mathbf{X}$ alongside a temporally sparse panoramic video $\mathbf{P}$. The panoramic stream captures scene evolution across all viewing directions, while the perspective stream adheres to the prescribed camera trajectory (see Fig.~\ref{fig:teaser}). Both streams share a single diffusion transformer backbone and exchange features via Geometric Anchoring (GeoAnch) and Temporal Compensation (TemComp), as illustrated in Fig.~\ref{fig:framework}. Both streams rely solely on the perspective prefix, text prompt, and camera parameters, 
without requiring any auxiliary depth, panoramic images, or video inputs.

\subsection{Coupled Panoramic Evolution}
\label{sec:joint}
We jointly denoise the dense perspective stream and the panoramic stream 
using a shared pretrained video diffusion transformer~\cite{wan2025wan}. 
Because predicting a full-rate panoramic stream incurs prohibitive computational overhead, we design the panoramic stream to operate at a lower frame rate. 
With a panoramic sampling interval $s$, panoramic frame $\mathbf{P}_j$ corresponds to perspective frame $\mathbf{X}_{sj}$ in a $T$-frame video, yielding $J=(T-1)/s+1$ states over the same temporal horizon.
These states can represent scene dynamics across all viewing directions around the camera, including regions outside the perspective viewport.

\noindent\textbf{Joint Denoising of Perspective and Panoramic Streams.}
For each stream, the noisy latents and corresponding VAE-encoded~\cite{kingma2014auto} visual conditions are concatenated channel-wise and mapped into token sequences via view-specific input projections.
The resulting token sequences from both perspective and panoramic streams are then concatenated along the sequence dimension.
Joint spatiotemporal self-attention subsequently processes both streams under a shared text prompt and diffusion timestep, after which a shared prediction head outputs the flow fields to be split and reshaped back into their respective latent grids. 
Retaining per-view spatial RoPE~\cite{su2024roformer}, we assign temporal RoPE coordinates $i$ and $sj$ 
to perspective and panoramic latent frames $i$ and $j$ (corresponding to original frame timestamps $4i$ and $4sj$ 
under $4\times$ temporal VAE compression), aligning synchronized states.

\noindent\textbf{Visual Conditions.}
We construct two visual conditions from the observed perspective prefix $\mathbf{X}_{0:K-1}$. 
The perspective condition comprises the VAE encoding of the prefix concatenated with an indicator mask denoting known frames. 
To construct the panoramic condition, the observed frames are reprojected onto  equirectangular projection (ERP) coordinates using camera poses and intrinsics, encoded via the VAE, and concatenated with a spatial visibility mask. 
Unobserved future frames and unmapped ERP regions are filled with zeros prior to VAE encoding. 
Crucially, both condition tensors are derived from the single perspective prefix, requiring no auxiliary visual inputs.

\noindent\textbf{Camera Conditions.}
We align the initial panoramic coordinate frame with the first perspective view (centered at the ERP center), 
so both cameras share the same initial orientation. 
Representing camera-to-world poses as $\mathbf{C}_t^{\mathrm{x}},\mathbf{C}_j^{\mathrm{p}}\in\mathbb{R}^{4\times4}$ 
and the pure-rotation viewport transformation from the perspective to the panoramic frame as $\mathbf{V}_t\in\mathbb{R}^{4\times4}$ (with $\mathbf{V}_0=\mathbf{I}$), 
the panoramic pose is obtained as $\mathbf{C}_j^{\mathrm{p}}=\mathbf{C}_{sj}^{\mathrm{x}}\mathbf{V}_{sj}^{-1}$. 
This operation eliminates viewport rotation while preserving shared camera centers, 
yielding $J=(T-1)/s+1$ panoramic poses when subsampling every $s$ frames. 
During training, $\mathbf{V}_t$ is known from the viewport rotations used to render perspective clips from source panoramas. 
At inference, the panoramic camera retains the initial perspective orientation $\mathbf{R}_0^{\mathrm{x}}\in\mathbb{R}^{3\times3}$, 
setting the rotation component of $\mathbf{V}_t$ to $(\mathbf{R}_0^{\mathrm{x}})^\top\mathbf{R}_t^{\mathrm{x}}$. 
Consequently, all required poses are determined solely by the input perspective trajectory without additional camera inputs.

\subsection{Geometric Anchoring}
\label{sec:geoanch}
Because spatial RoPE~\cite{su2024roformer} in diffusion transformers operates independently on separate image grids, it does not explicitly capture geometric correspondences between perspective and panoramic views. 
Although spherical positional embeddings, such as SpheRoPE~\cite{hirschorn2026spherope}, offer an alternative formulation, directly replacing pretrained spatial RoPE with spherical coordinates disrupts learned spatial priors and impedes joint optimization. 
We therefore introduce Geometric Anchoring (GeoAnch) to establish explicit cross-view correspondences while preserving the pretrained spatial encoding.

Hereafter, $i$ and $j$ index perspective and panoramic latent frames (corresponding to RGB frames $4i$ and $4j$), respectively. 
Let $\mathbf{q}$ denote the image-space patch center of a perspective token, and let $\mathbf{A}_i\in\mathbb{R}^{3\times3}$, $\mathbf{R}_i^{\mathrm{x}}\in\mathbb{R}^{3\times3}$, and $\mathbf{R}_j^{\mathrm{p}}\in\mathbb{R}^{3\times3}$ denote the corresponding camera intrinsics and rotations. 
At synchronized timestamps ($i=sj$), both views share an optical center, yielding the mapping:
\begin{equation}
 \pi_{i\to j}(\mathbf{q})=\Pi_{\mathrm{erp}}\!\left((\mathbf{R}_j^{\mathrm{p}})^\top \mathbf{R}_i^{\mathrm{x}} \mathbf{A}_i^{-1}\widetilde{\mathbf{q}}\right),
 \label{eq:projection}
\end{equation}
where $\widetilde{\mathbf{q}}$ is the homogeneous image coordinate and $\Pi_{\mathrm{erp}}$ projects a 3D ray onto the panoramic grid.
Conversely, panoramic directions that fall within the perspective field of view are mapped back to the perspective grid using the inverse geometric transformation.

Based on these geometric correspondences, GeoAnch bilinearly samples aligned features from the other view, concatenates them with the current view's features, and estimates a residual update using a learned linear layer. 
Because the perspective camera observes only a sub-region of the visual sphere, updates to the panoramic stream are masked to the perspective field of view. 
In addition, feature sampling on the ERP grid employs horizontal circular wrapping to preserve spherical boundary continuity.

\subsection{Temporal Compensation}
\label{sec:temcomp}

Due to the temporal sparsity of the panoramic stream, intermediate perspective frames lack synchronized panoramic anchors. 
Temporal Compensation (TemComp) bridges this gap through bidirectional cross-temporal feature aggregation. 
For each intermediate perspective frame, TemComp performs interpolation between its two adjacent panoramic states. 
Conversely, each panoramic frame aggregates a local temporal window of perspective features centered at its synchronized timestamp 
to learn a correction relative to its synchronized geometric anchor.

Let $\mathbf{g}$ denote a query patch feature representation of the current frame. 
Following the ray projection in Eq.~(\ref{eq:projection}), we map each patch location of the current frame into candidate frames of the other view, bilinearly sampling features to obtain $\widetilde{\mathbf{h}}_k$ at candidate timestamp $k$. 
Candidate features are aggregated as:
\begin{equation}
 \bar{\mathbf{h}}(\mathbf{g}) \! = \! \sum_{k\in\mathcal{N}(\mathbf{g})} \! \! \alpha_k\widetilde{\mathbf{h}}_k, \! \! \quad
 \alpha_k \propto  \exp\!\left(\operatorname{cos}(\mathbf{g},\widetilde{\mathbf{h}}_k) / \theta\right),
 \label{eq:soft}
\end{equation}
where $\mathcal{N}(\mathbf{g})$ spans two bracketing panoramic frames for perspective queries, or up to $2\lfloor s/2\rfloor+1$ centered perspective frames for panoramic queries.
The aggregation weights $\alpha_k$ are normalized to sum to one, and $\theta=0.2$ serves as the temperature hyperparameter. 
The geometric mapping provides approximate directional correspondences to guide candidate sampling, while feature similarity suppresses incompatible evidence across time.

To update representations, TemComp incorporates a learned linear residual layer for each exchange direction, preserving base features while attenuating unreliable contributions. 
Specifically, synchronized perspective frames retain their direct GeoAnch updates, whereas intermediate perspective frames receive a residual computed from the aggregated representation $\bar{\mathbf{h}}$. 
TemComp is thus formulated as a conservative local feature correction around synchronized geometric anchors, enabling the model to selectively exploit neighboring temporal observations.

\subsection{Training and Inference}
\label{sec:learning}

Perspective frames are directly encoded using the pretrained VAE~\cite{kingma2014auto}. 
For panoramic targets, we adopt circular latent encoding~\cite{wu2026360anything}, which applies horizontal circular padding to ERP frames prior to VAE encoding and subsequently crops the padded latent margins to restore the original grid dimensions. 
This provides a seamless latent representation across horizontal boundaries for diffusion training.

Flow matching and loss optimization operate directly on these latent spaces. 
For each view $v\in\{\mathrm{x},\mathrm{p}\}$, we sample independent Gaussian noise $\boldsymbol{\epsilon}^v\sim\mathcal{N}(\mathbf{0},\mathbf{I})$ 
at a shared noise level $\sigma\in[0,1]$ to define the linear interpolation and velocity target:
\begin{equation}
 \mathbf{z}_\sigma^v=(1-\sigma)\mathbf{z}^v+\sigma\boldsymbol{\epsilon}^v,
 \qquad \mathbf{u}^v=\boldsymbol{\epsilon}^v-\mathbf{z}^v.
 \label{eq:flow}
\end{equation}
The diffusion transformer jointly predicts the velocity fields $\hat{\mathbf{u}}^{\mathrm{x}}$ and $\hat{\mathbf{u}}^{\mathrm{p}}$. 
We train the model by minimizing the weighted objective:
\begin{equation}
 \mathcal{L} \!  = \! \mathbb{E}_{\sigma,\boldsymbol{\epsilon}}\!\left[\frac{w(\sigma)}{2}\left(\|\hat{\mathbf{u}}^{\mathrm{x}} \!\! - \!\! \mathbf{u}^{\mathrm{x}}\|^2 \! + \!  \|\mathbf{W}_{\mathrm{p}}\odot(\hat{\mathbf{u}}^{\mathrm{p}} \!\! - \!\! \mathbf{u}^{\mathrm{p}})\|^2\right)\right],
 \label{eq:loss}
\end{equation}
where $w(\sigma)$ is the scheduler weight and $\mathbf{W}_{\mathrm{p}}$ denotes a normalized, latitude-dependent loss weight for the panoramic canvas. 
During training, the LoRA~\cite{hu2022lora} parameters, input projections, camera conditioning modules, and cross-view interaction layers are optimized jointly, while the VAE and text encoder remain frozen.

During inference, the shared VAE can decode both streams. 
When panoramic visualization is not explicitly required, only the perspective stream is decoded, incurring zero additional decoding overhead.
\section{Experiments}
\label{sec:experiments}

\begin{table*}[t!]
\centering
\caption{\textbf{Image-to-video (I2V) comparison.} $\dagger$ indicates training on our dataset with the same Wan2.2-TI2V-5B backbone and LoRA settings. 
Unmarked methods use released checkpoints directly, with parameter counts of diffusion models in parentheses.}
\vspace{-2mm}
\label{tab:main-i2v}
\scriptsize
\setlength{\tabcolsep}{2pt}
\resizebox{\textwidth}{!}{%
\begin{tabular}{@{}lcccccccccccccc@{}}
\toprule
\multirow{2}{*}{Methods} & \multicolumn{3}{c}{Overall Quality} & \multicolumn{4}{c}{Camera Controllability} & \multicolumn{3}{c}{BG Revisit Fidelity} & \multicolumn{4}{c}{Dynamic Subject Persistence} \\
\cmidrule(lr){2-4}\cmidrule(lr){5-8}\cmidrule(lr){9-11}\cmidrule(lr){12-15}
 & FID\dn & FVD\dn & FVD$_c$\dn & Rot\dn & Trans\dn & CamMC\dn & ATE\dn & PSNR\up & SSIM\up & LPIPS\dn & ID\up & R-ID\up & Vis.\up & Traj.\dn \\
\midrule
HyDRA (1.3B)~\cite{chen2026hybrid} & 48.51 & 857.4 & 763.8 & 35.949 & 0.707 & 1.152 & 0.805 & 11.09 & 0.3986 & 0.676 & \secondcell{0.6725} & 0.1319 & \secondcell{0.6376} & \secondcell{0.4876} \\
Matrix-World 3.0 (5B)~\cite{wang2026matrix} & 41.18 & 688.6 & 654.9 & 28.952 & 0.663 & 1.008 & 0.748 & 9.74 & 0.3747 & 0.693 & 0.5570 & 0.1643 & 0.5739 & 0.6456 \\
DreamX-World (5B)~\cite{team2026dreamx} & \thirdcell{26.28} & \thirdcell{392.3} & \secondcell{368.4} & 24.743 & \thirdcell{0.569} & 0.876 & \thirdcell{0.636} & 11.78 & 0.4412 & \thirdcell{0.638} & 0.6044 & \thirdcell{0.2075} & 0.5916 & 0.5621 \\
SANA-WM (2.6B+17B)~\cite{zhu2026sana} & 33.28 & 432.9 & 402.4 & \thirdcell{20.116} & 0.590 & \thirdcell{0.819} & 0.655 & \thirdcell{12.22} & \thirdcell{0.4488} & 0.644 & \thirdcell{0.6108} & 0.1278 & 0.6108 & 0.5262 \\
LiveWorld (14B)~\cite{duan2026liveworld} & \secondcell{24.53} & \firstcell{335.4} & \firstcell{334.4} & \secondcell{16.252} & \secondcell{0.439} & \secondcell{0.655} & \secondcell{0.492} & \firstcell{12.46} & \firstcell{0.4664} & \firstcell{0.595} & 0.6035 & \secondcell{0.2928} & \thirdcell{0.6229} & \thirdcell{0.5016} \\
LingBot-World 2.0 (28B)~\cite{gao2026infinite} & \firstcell{22.64} & \secondcell{383.8} & \thirdcell{389.4} & \firstcell{13.560} & \firstcell{0.422} & \firstcell{0.587} & \firstcell{0.470} & \secondcell{12.42} & \secondcell{0.4645} & \secondcell{0.609} & \firstcell{0.6964} & \firstcell{0.3360} & \firstcell{0.7303} & \firstcell{0.3853} \\
\midrule
CameraCtrl\textsuperscript{\dag}~\cite{he2025cameractrl} & 23.60 & 288.5 & 295.1 & 13.086 & 0.531 & 0.677 & 0.594 & 12.29 & 0.4415 & 0.618 & 0.5834 & 0.2325 & 0.6519 & 0.4753 \\
~\cite{he2025cameractrl}+OmniPersist ($s=1$) & \thirdcell{22.64} & \secondcell{258.6} & \thirdcell{266.8} & 6.145 & 0.422 & 0.490 & 0.470 & \thirdcell{13.47} & \secondcell{0.4749} & \secondcell{0.575} & \firstcell{0.6525} & 0.2653 & \firstcell{0.7375} & \secondcell{0.3779} \\
~\cite{he2025cameractrl}+OmniPersist ($s=2$) & \secondcell{22.54} & \firstcell{253.8} & \secondcell{256.0} & \secondcell{5.586} & \secondcell{0.415} & \secondcell{0.474} & \secondcell{0.463} & \secondcell{13.52} & \secondcell{0.4749} & \thirdcell{0.576} & \thirdcell{0.6411} & \secondcell{0.2933} & \secondcell{0.7340} & \firstcell{0.3759} \\
~\cite{he2025cameractrl}+OmniPersist ($s=4$) & \firstcell{22.18} & \thirdcell{262.3} & \firstcell{252.6} & \firstcell{5.556} & \firstcell{0.399} & \firstcell{0.458} & \firstcell{0.446} & \firstcell{13.58} & \firstcell{0.4754} & \firstcell{0.573} & 0.6395 & \firstcell{0.3355} & 0.7290 & 0.3830 \\
~\cite{he2025cameractrl}+OmniPersist ($s=10$) & 22.91 & 266.6 & 276.4 & \thirdcell{5.695} & \thirdcell{0.418} & \thirdcell{0.478} & \thirdcell{0.465} & 13.37 & \thirdcell{0.4731} & 0.580 & \secondcell{0.6429} & \thirdcell{0.2840} & \thirdcell{0.7333} & \thirdcell{0.3783} \\
\midrule
UCPE\textsuperscript{\dag}~\cite{zhang2026ucpe} & 25.33 & 333.7 & 325.7 & 8.127 & 0.431 & 0.526 & 0.481 & 12.80 & 0.4696 & 0.580 & 0.6043 & 0.2097 & 0.6561 & 0.4897 \\

~\cite{zhang2026ucpe}+OmniPersist ($s=1$) & \firstcell{21.41} & \thirdcell{274.1} & \secondcell{262.2} & \firstcell{4.236} & 0.382 & \secondcell{0.426} & 0.427 & \secondcell{13.91} & \secondcell{0.4890} & \firstcell{0.543} & \secondcell{0.6435} & \thirdcell{0.3183} & \thirdcell{0.7348} & \thirdcell{0.4003} \\
~\cite{zhang2026ucpe}+OmniPersist ($s=2$) & \thirdcell{21.90} & \firstcell{264.6} & \firstcell{261.5} & \secondcell{4.490} & \secondcell{0.374} & \firstcell{0.425} & \secondcell{0.418} & \firstcell{13.99} & \firstcell{0.4898} & \firstcell{0.543} & \thirdcell{0.6416} & \firstcell{0.3839} & \firstcell{0.7412} & \secondcell{0.3828} \\
~\cite{zhang2026ucpe}+OmniPersist ($s=4$)  & \secondcell{21.89} & \secondcell{272.7} & \thirdcell{264.0} & \thirdcell{4.976} & \thirdcell{0.376} & \thirdcell{0.429} & \thirdcell{0.422} & \thirdcell{13.90} & \thirdcell{0.4872} & \secondcell{0.546} & \firstcell{0.6486} & \secondcell{0.3694} & \secondcell{0.7351} & \firstcell{0.3791} \\
~\cite{zhang2026ucpe}+OmniPersist ($s=10$) & 22.57 & 285.8 & 284.7 & 5.670 & \firstcell{0.365} & 0.430 & \firstcell{0.408} & 13.76 & 0.4825 & \thirdcell{0.553} & 0.6283 & 0.3116 & 0.7338 & 0.4090 \\
\bottomrule
\end{tabular}%
}
\end{table*}

\begin{table*}[t!]
\centering
\caption{\textbf{Video continuation comparison.} $\dagger$ indicates training on our dataset with the same Wan2.2-TI2V-5B backbone and LoRA settings. 
Unmarked methods use released checkpoints directly, with parameter counts of diffusion models in parentheses.}
\vspace{-2mm}
\label{tab:main-continuation}
\scriptsize
\setlength{\tabcolsep}{2pt}
\resizebox{\textwidth}{!}{%
\begin{tabular}{@{}lcccccccccccccc@{}}
\toprule
\multirow{2}{*}{Methods} & \multicolumn{3}{c}{Overall Quality} & \multicolumn{4}{c}{Camera Controllability} & \multicolumn{3}{c}{BG Revisit Fidelity} & \multicolumn{4}{c}{Dynamic Subject Persistence} \\
\cmidrule(lr){2-4}\cmidrule(lr){5-8}\cmidrule(lr){9-11}\cmidrule(lr){12-15}
 & FID\dn & FVD\dn & FVD$_c$\dn & Rot\dn & Trans\dn & CamMC\dn & ATE\dn & PSNR\up & SSIM\up & LPIPS\dn & ID\up & R-ID\up & Vis.\up & Traj.\dn \\
\midrule
HyDRA (1.3B)~\cite{chen2026hybrid} & 61.30 & 779.7 & 711.2 & 41.420 & 0.724 & 1.259 & 0.787 & \thirdcell{11.30} & 0.3955 & 0.674 & 0.3102 & 0.0850 & 0.3816 & \thirdcell{0.7474} \\
Matrix-World 3.0 (5B)~\cite{wang2026matrix} & \thirdcell{36.94} & \thirdcell{427.2} & \thirdcell{426.8} & \thirdcell{25.654} & \thirdcell{0.588} & \thirdcell{0.922} & \thirdcell{0.644} & 10.86 & \thirdcell{0.4040} & \thirdcell{0.666} & \thirdcell{0.3386} & \thirdcell{0.1353} & \thirdcell{0.4229} & 0.7710 \\
SANA-WM (2.6B+17B)~\cite{zhu2026sana} & \secondcell{32.75} & \secondcell{320.8} & \secondcell{330.6} & \secondcell{22.923} & \secondcell{0.546} & \secondcell{0.855} & \secondcell{0.588} & \secondcell{12.86} & \secondcell{0.4578} & \secondcell{0.606} & \secondcell{0.4019} & \firstcell{0.2039} & \firstcell{0.5006} & \secondcell{0.6568} \\
LiveWorld (14B)~\cite{duan2026liveworld} & \firstcell{25.45} & \firstcell{286.7} & \firstcell{295.0} & \firstcell{12.912} & \firstcell{0.425} & \firstcell{0.605} & \firstcell{0.461} & \firstcell{13.26} & \firstcell{0.4797} & \firstcell{0.556} & \firstcell{0.4325} & \secondcell{0.1762} & \secondcell{0.4824} & \firstcell{0.6358} \\
\midrule
CameraCtrl\textsuperscript{\dag}~\cite{he2025cameractrl} & 22.61 & 227.5 & 223.5 & 11.704 & \thirdcell{0.451} & 0.597 & 0.489 & 13.43 & 0.4701 & 0.579 & 0.4946 & 0.2494 & 0.6053 & 0.5161 \\
~\cite{he2025cameractrl}+OmniPersist ($s=1$) & 21.87 & \thirdcell{197.2} & \secondcell{201.8} & 4.831 & \secondcell{0.403} & 0.459 & \thirdcell{0.439} & \secondcell{15.28} & \secondcell{0.5245} & \firstcell{0.511} & \firstcell{0.5667} & \firstcell{0.4093} & \firstcell{0.7042} & \firstcell{0.4117} \\
~\cite{he2025cameractrl}+OmniPersist ($s=2$) & \thirdcell{21.70} & \secondcell{196.6} & 202.9 & \thirdcell{4.690} & \firstcell{0.401} & \thirdcell{0.456} & \firstcell{0.433} & \firstcell{15.31} & \firstcell{0.5263} & \secondcell{0.514} & \secondcell{0.5655} & \secondcell{0.4062} & \thirdcell{0.6768} & \secondcell{0.4177} \\
~\cite{he2025cameractrl}+OmniPersist ($s=4$) & \firstcell{21.59} & \firstcell{194.9} & \thirdcell{201.9} & \firstcell{4.345} & \firstcell{0.401} & \firstcell{0.450} & \firstcell{0.433} & \thirdcell{15.22} & \thirdcell{0.5238} & \thirdcell{0.515} & 0.5534 & \thirdcell{0.3771} & 0.6678 & 0.4289 \\
~\cite{he2025cameractrl}+OmniPersist ($s=10$) & \secondcell{21.69} & 201.0 & \firstcell{200.6} & \secondcell{4.661} & \firstcell{0.401} & \secondcell{0.455} & \secondcell{0.435} & 15.20 & 0.5221 & 0.518 & \thirdcell{0.5559} & 0.3492 & \secondcell{0.6835} & \thirdcell{0.4287} \\
\midrule
UCPE\textsuperscript{\dag}~\cite{zhang2026ucpe} & 23.60 & 264.6 & 254.7 & 6.711 & 0.443 & 0.519 & 0.476 & 14.55 & 0.5136 & 0.523 & 0.4932 & 0.2778 & 0.6354 & 0.5255 \\
~\cite{zhang2026ucpe}+OmniPersist ($s=1$) & \firstcell{20.53} & \firstcell{196.7} & \firstcell{197.9} & \firstcell{3.265} & \firstcell{0.338} & \firstcell{0.375} & \firstcell{0.367} & \firstcell{15.90} & \firstcell{0.5446} & \firstcell{0.478} & \thirdcell{0.5504} & 0.3669 & \thirdcell{0.6759} & \thirdcell{0.4280} \\
~\cite{zhang2026ucpe}+OmniPersist ($s=2$) & \secondcell{21.04} & \secondcell{197.6} & \secondcell{202.4} & \secondcell{3.863} & \secondcell{0.341} & \secondcell{0.388} & \secondcell{0.369} & \secondcell{15.79} & \secondcell{0.5406} & \secondcell{0.484} & \firstcell{0.5580} & \firstcell{0.4667} & \firstcell{0.6829} & \firstcell{0.4173} \\
~\cite{zhang2026ucpe}+OmniPersist ($s=4$) & \thirdcell{21.41} & \thirdcell{210.3} & \thirdcell{212.6} & \thirdcell{4.169} & \thirdcell{0.342} & \thirdcell{0.393} & \thirdcell{0.372} & \thirdcell{15.72} & \thirdcell{0.5385} & \secondcell{0.484} & \secondcell{0.5577} & \thirdcell{0.3849} & \secondcell{0.6800} & \secondcell{0.4245} \\
~\cite{zhang2026ucpe}+OmniPersist ($s=10$) & 22.48 & 218.1 & 219.1 & 4.353 & 0.354 & 0.407 & 0.384 & 15.39 & 0.5234 & \thirdcell{0.496} & 0.5126 & \secondcell{0.4045} & 0.6649 & 0.4618 \\
\bottomrule
\end{tabular}%
}
\vspace{0mm}
\end{table*}

\subsection{Experimental Setup}
\noindent\textbf{Dataset Construction.}
Unlike HyDRA~\cite{chen2026hybrid}, which constructs its training and evaluation sets within simulators (\ie, Unreal Engine 5) that exhibit limited visual realism, we curate our dataset directly from real-world environments. 
We construct paired panoramic and perspective videos by simulating camera viewing trajectories over panoramic footage from the 360-1M dataset~\cite{wallingford2024image}. 
Following the filtering protocols in Argus~\cite{luo2025beyond} and UCPE~\cite{zhang2026ucpe}, $\sim$125K high-quality video clips are selected. 
Each clip consists of 81 frames at 16 fps, with a perspective resolution of $832\times480$ and a panoramic resolution of $1024\times512$. 
We design four distinct camera trajectory types: object tracking, single revisit, double revisit, and a full camera rotation of $\sim$360$^\circ$, covering both continuous subject following and leave-and-revisit patterns. 
Perspective viewpoints retain the camera-center trajectory of the original panoramic video while dynamically varying the viewing orientation. 
The evaluation set contains 300 videos whose source footage is strictly excluded from the training split. 
Details of the construction pipeline are provided in the \textit{suppl}.

\noindent\textbf{Implementation Details.}
Under the image-to-video (I2V) protocol, generation is conditioned on a single perspective frame ($K=1$), whereas the video continuation setting conditions on 21 frames ($K=21$). 
Both settings target the generation of 81-frame videos. 
OmniPersist is implemented based on the Wan2.2-TI2V-5B backbone~\cite{wan2025wan}. 
We instantiate OmniPersist across two representative camera-control paradigms, applying rank-32 LoRA~\cite{hu2022lora} to linear layers in the self-attention and feed-forward blocks. 
Under CameraCtrl~\cite{he2025cameractrl}, view-specific ray adapters inject the camera features of the two views into their respective input tokens. 
Under UCPE~\cite{zhang2026ucpe}, camera-attention layers process both views according to their ray geometries under a unified coordinate and scale reference. 
GeoAnch and TemComp are embedded after each DiT Block.
All OmniPersist variants adopt mixed-interval sampling, uniformly drawing $s\in\{1,2,4,10\}$ and prefix length $K\in\{1,21\}$ per example. 
Using AdamW~\cite{loshchilov2018decoupled} with a learning rate of $10^{-4}$ and total batch size of 8 on 8 NVIDIA RTX PRO 6000 GPUs, the UCPE- and CameraCtrl-based variants converge in 20K and 80K iterations, respectively.

\noindent\textbf{Comparison Setup.}
We benchmark against world models developed through large-scale, dedicated pretraining, \ie, DreamX-World~\cite{team2026dreamx}, Matrix-World 3.0~\cite{wang2026matrix}, SANA-WM~\cite{zhu2026sana}, and LingBot-World 2.0~\cite{gao2026infinite}. 
We evaluate their released checkpoints configured with bidirectional attention.
Beyond these, camera-conditioned video generation serves as a foundational paradigm for world models. 
To ensure a fair comparison, we train CameraCtrl~\cite{he2025cameractrl} and UCPE~\cite{zhang2026ucpe} using the same dataset, training settings, Wan2.2-TI2V-5B backbone, and LoRA configuration as OmniPersist.
Furthermore, we include comparisons with HyDRA~\cite{chen2026hybrid} and LiveWorld~\cite{duan2026liveworld}, which are specifically designed for out-of-sight dynamic persistence. 
During inference, all diffusion models employ 50 denoising steps. 
Some models are evaluated solely on I2V, due to incompatibility with video continuation tasks.

\noindent\textbf{Evaluation Protocol.}
We evaluate performance across 14 perspective-video metrics structured into four groups. 
(1) Overall visual quality is measured via FID~\cite{heusel2017fid}, FVD~\cite{unterthiner2018fvd}, and center-crop FVD (FVD$_c$). 
(2) Camera controllability is evaluated via rotation (Rot) and translation (Trans) errors~\cite{he2025cameractrl}, 
CamMC~\cite{wang2024motionctrl}, and absolute trajectory error (ATE) on ViPE-reconstructed~\cite{huang2025vipe} poses anchored at frame zero. 
Following UCPE~\cite{zhang2026ucpe}, predicted and reference translations are independently normalized by their maximum trajectory radius.
(3) Background revisit fidelity is evaluated using PSNR, SSIM~\cite{wang2004ssim}, and LPIPS~\cite{zhang2018unreasonable} between generated and ground-truth frames at camera-selected revisit timestamps, masked by intersected static backgrounds. 
(4) Dynamic persistence is assessed via foreground DINOv2~\cite{oquab2023dinov2} feature similarity for overall subject identity (ID) and re-entry identity (R-ID), visibility F1-score (Vis.), and normalized subject trajectory error (Traj.). These metrics evaluate whether a dynamic subject successfully returns, retains its appearance, and follows the ground-truth trajectory (the latter being more informative for video continuation).
Detailed metric implementations are provided in the \textit{suppl}.

\begin{figure*}[t!]
    \centering
    \includegraphics[width=\textwidth]{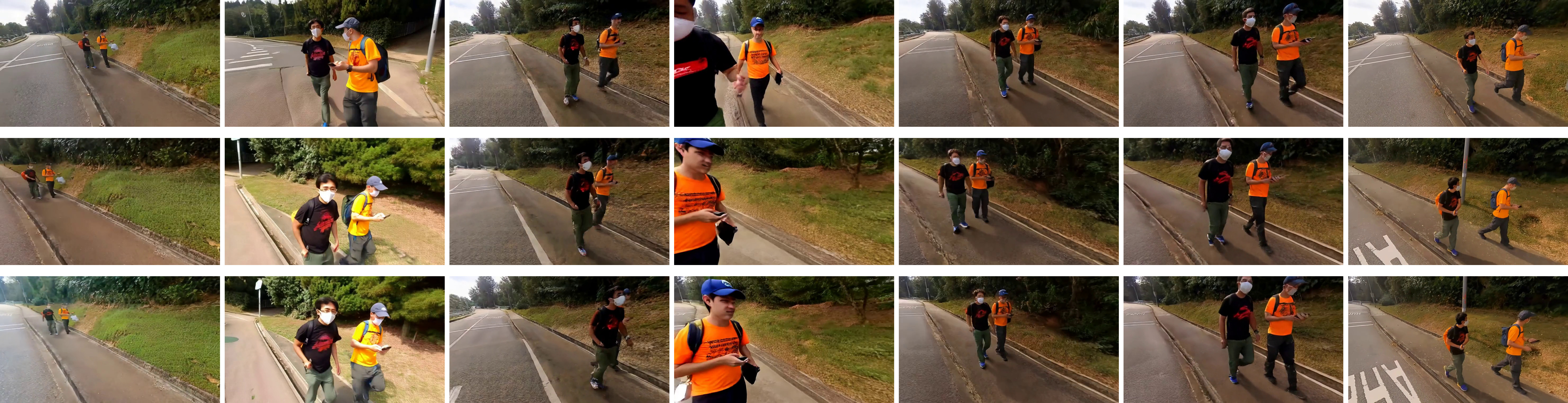}\par
    \vspace{-1mm}
    {
        \fontsize{7}{7}\selectfont
        \setlength{\tabcolsep}{0pt}
        \begin{tabular}{@{}*{7}{>{\centering\arraybackslash}p{\dimexpr\textwidth/7\relax}}@{}}
        LiveWorld & LingBot-World 2.0 & CameraCtrl & UCPE & CameraCtrl+OmniPersist & UCPE+OmniPersist & Ground Truth \\
        \end{tabular}
    }
    \vspace{-6mm}
    \caption{\textbf{Generation results on I2V across camera revisits.} 
    From top to bottom, we show a frame before the camera turns away and two revisit frames after it returns to illustrate identity preservation and continued motion. 
    Frames in each row share the same frame index.
    Prompt overview: \textit{a video shows two men walking along a paved path next to a road.}}
    \vspace{-3mm}
    \label{fig:qualitative-comparisons}
\end{figure*}

\subsection{Quantitative and Qualitative Results}
We adopt $s=4$ as the primary setting for both quantitative and qualitative evaluations. 

\noindent\textbf{Quantitative Comparisons.}
As shown in Table~\ref{tab:main-i2v}, under the I2V task, \method{} consistently improves all 14 metrics over the respective baselines for both CameraCtrl~\cite{he2025cameractrl} and UCPE~\cite{zhang2026ucpe}. 
Specifically, re-entry identity increases from 0.2325 to 0.3355 with CameraCtrl and from 0.2097 to 0.3694 with UCPE. 
Camera rotation and translation errors drop from 13.086 / 0.531 to 5.556 / 0.399 for CameraCtrl, and from 8.127 / 0.431 to 4.976 / 0.376 for UCPE. FVD under UCPE improves from 333.7 to 272.7. 
Notably, even when compared against larger released world models, including LiveWorld~\cite{duan2026liveworld} (14B, which explicitly models multi-entity evolution by additional diffusion models) and LingBot-World 2.0~\cite{gao2026infinite} (28B), the UCPE-based \method{} outperforms LiveWorld across all 14 metrics and surpasses LingBot-World 2.0 on 13 of the 14 metrics. 

As reported in Table~\ref{tab:main-continuation}, under the video continuation task, both variants again surpass their respective baselines across all 14 metrics. 
In particular, re-entry identity increases from 0.2494 to 0.3771 with CameraCtrl and from 0.2778 to 0.3849 with UCPE. 
Subject trajectory error decreases from 0.5161 to 0.4289 with CameraCtrl and from 0.5255 to 0.4245 with UCPE. 
Moreover, CameraCtrl-based \method{} improves background revisit PSNR by 1.79~dB and camera rotation error by 7.359. 
Both variants also outperform HyDRA~\cite{chen2026hybrid}, LiveWorld~\cite{duan2026liveworld}, Matrix-World 3.0~\cite{wang2026matrix}, and SANA-WM~\cite{zhu2026sana} across all 14 metrics. 

In summary, \method{} achieves consistent gains in overall visual quality, camera controllability, background revisit fidelity, and dynamic subject persistence across both generation tasks and camera control mechanisms.

\noindent\textbf{Qualitative Comparisons.}
\label{sec:qualitative}
Figure~\ref{fig:qualitative-comparisons} demonstrates that \method{} maintains strong consistency for both static backgrounds and dynamic subjects across a camera revisit. 
Specifically, the road layout and surrounding vegetation remain structurally coherent, while both pedestrians preserve their clothing details and recognizable appearances upon re-entry. 
Compared with perspective-only baselines, our results better preserve both subjects and their continued walking motion after the camera returns. 
More qualitative results and panoramic visualization are provided in the \textit{suppl}.

\begin{table*}[t!]
\centering
\caption{\textbf{Component ablation across both tasks.} The evaluations are conducted using CameraCtrl~\cite{he2025cameractrl} on $s=4$.}
\vspace{-3mm}
\label{tab:component-camera}
\scriptsize
\setlength{\tabcolsep}{2pt}
\resizebox{\textwidth}{!}{%
\begin{tabular}{@{}lcccccccccccccc@{}}
\toprule
\multirow{2}{*}{Configuration} & \multicolumn{3}{c}{Overall Quality} & \multicolumn{4}{c}{Camera Controllability} & \multicolumn{3}{c}{BG Revisit Fidelity} & \multicolumn{4}{c}{Dynamic Subject Persistence} \\
\cmidrule(lr){2-4}\cmidrule(lr){5-8}\cmidrule(lr){9-11}\cmidrule(lr){12-15}
 & FID\dn & FVD\dn & FVD$_c$\dn & Rot\dn & Trans\dn & CamMC\dn & ATE\dn & PSNR\up & SSIM\up & LPIPS\dn & ID\up & R-ID\up & Vis.\up & Traj.\dn \\
\midrule
\multicolumn{15}{l}{\cellcolor{gray!15}\textit{Image-to-video ($K=1$)}} \\
w/o Pano & \thirdcell{23.60} & 288.5 & 295.1 & 13.086 & 0.531 & 0.677 & 0.594 & 12.29 & 0.4415 & 0.618 & 0.5834 & 0.2325 & 0.6519 & 0.4753 \\
w/ Pano & 26.16 & \thirdcell{288.1} & \thirdcell{281.0} & \thirdcell{9.000} & \thirdcell{0.467} & \thirdcell{0.562} & \thirdcell{0.519} & \thirdcell{12.69} & \thirdcell{0.4473} & \thirdcell{0.612} & \thirdcell{0.5975} & \secondcell{0.3256} & \thirdcell{0.6700} & \thirdcell{0.4645} \\
w/ Pano \& GeoAnch & \secondcell{23.07} & \secondcell{275.2} & \secondcell{265.9} & \secondcell{6.604} & \secondcell{0.438} & \secondcell{0.508} & \secondcell{0.493} & \firstcell{13.63} & \firstcell{0.4817} & \secondcell{0.589} & \secondcell{0.6163} & \thirdcell{0.2794} & \secondcell{0.7245} & \secondcell{0.3960} \\
w/ Pano \& GeoAnch \& TemComp & \firstcell{22.18} & \firstcell{262.3} & \firstcell{252.6} & \firstcell{5.556} & \firstcell{0.399} & \firstcell{0.458} & \firstcell{0.446} & \secondcell{13.58} & \secondcell{0.4754} & \firstcell{0.573} & \firstcell{0.6395} & \firstcell{0.3355} & \firstcell{0.7290} & \firstcell{0.3830} \\
\midrule
\multicolumn{15}{l}{\cellcolor{gray!15}\textit{Video continuation ($K=21$)}} \\
w/o Pano & \secondcell{22.61} & \secondcell{227.5} & \secondcell{223.5} & 11.704 & \thirdcell{0.451} & 0.597 & \thirdcell{0.489} & 13.43 & \thirdcell{0.4701} & \thirdcell{0.579} & \thirdcell{0.4946} & 0.2494 & 0.6053 & \thirdcell{0.5161} \\
w/ Pano & 27.20 & 246.0 & 232.7 & \thirdcell{8.453} & 0.462 & \thirdcell{0.560} & 0.497 & \thirdcell{13.64} & 0.4682 & 0.581 & 0.4890 & \thirdcell{0.3167} & \thirdcell{0.6104} & 0.5201 \\
w/ Pano \& GeoAnch & \thirdcell{23.89} & \thirdcell{228.5} & \thirdcell{227.0} & \secondcell{5.905} & \secondcell{0.424} & \secondcell{0.493} & \secondcell{0.456} & \secondcell{14.82} & \secondcell{0.5162} & \secondcell{0.547} & \secondcell{0.5007} & \secondcell{0.3173} & \secondcell{0.6465} & \secondcell{0.4956} \\
w/ Pano \& GeoAnch \& TemComp & \firstcell{21.59} & \firstcell{194.9} & \firstcell{201.9} & \firstcell{4.345} & \firstcell{0.401} & \firstcell{0.450} & \firstcell{0.433} & \firstcell{15.22} & \firstcell{0.5238} & \firstcell{0.515} & \firstcell{0.5534} & \firstcell{0.3771} & \firstcell{0.6678} & \firstcell{0.4289} \\
\bottomrule
\end{tabular}%
}
\vspace{-4mm}
\end{table*}

\begin{table}[t]
\centering
\caption{Per-step denoising cost. 
Each entry reports FLOPs (T) / time (ms) / peak allocated memory (GB). The first and second rows correspond to CameraCtrl~\cite{he2025cameractrl} and UCPE~\cite{zhang2026ucpe}, respectively.}
\vspace{-3mm}
\label{tab:flops}
\scriptsize
\setlength{\tabcolsep}{2pt}
\resizebox{\linewidth}{!}{%
\begin{tabular}{@{}*{5}{c}@{}}
\toprule
\multirow{2}{*}{Baseline} & \multicolumn{4}{c}{+OmniPersist} \\
\cmidrule(lr){2-5}
 & $s=10$ & $s=4$ & $s=2$ & $s=1$ \\
\midrule
102 / 420 / 11.2 & 128 / 547 / 12.2 & 155 / 670 / 12.3 & 203 / 838 / 12.5 & 313 / 1185 / 13.1 \\
103 / 442 / 10.2 & 131 / 571 / 10.9 & 159 / 689 / 11.0 & 209 / 865 / 11.2 & 326 / 1249 / 11.6 \\
\bottomrule
\end{tabular}%
}
\vspace{-4mm}
\end{table}

\subsection{Ablation Studies}
We first isolate the contributions of panoramic evolution, GeoAnch, and TemComp using CameraCtrl~\cite{he2025cameractrl} at $s=4$. 
We then analyze the effect of panoramic sampling density and evaluate the computational overhead of our framework.

\noindent\textbf{Effect of Joint Panoramic Evolution.}
As shown in Table~\ref{tab:component-camera}, compared to the perspective-only baseline (w/o Pano), adding panoramic support (w/ Pano) improves re-entry identity (R-ID) from 0.2325 to 0.3256 for I2V and from 0.2494 to 0.3167 for video continuation. 
However, visual quality degrades inconsistently without explicit cross-view alignment: I2V FID worsens from 23.60 to 26.16, while video continuation FVD increases from 227.5 to 246.0. 
When explicit spatial and temporal coupling are incorporated, these scores improve substantially to 22.18 and 194.9, respectively. 
These results indicate that although the panoramic stream supplies essential information about returning subjects, naive joint modeling alone is insufficient to maintain generation quality.

\noindent\textbf{Effect of GeoAnch and TemComp.}
Adding GeoAnch improves all four camera-control and three background-revisit metrics over panoramic modeling alone (w/ Pano) in both tasks (see Table~\ref{tab:component-camera}). 
For I2V, rotation error drops from 9.000 to 6.604 and background PSNR rises from 12.69 to 13.63~dB, while trajectory error decreases from 0.4645 to 0.3960 for I2V and from 0.5201 to 0.4956 for video continuation. 
However, R-ID remains inconsistent, dropping from 0.3256 to 0.2794 for I2V and changing marginally for continuation. These results confirm the value of GeoAnch, yet show that synchronized correspondence alone is insufficient for identity recovery after out-of-view intervals.

Incorporating TemComp further improves all visual quality, camera-control, and dynamic-persistence metrics over the GeoAnch-only variant across both tasks. 
Specifically, R-ID rises from 0.2794 to 0.3355 for I2V and from 0.3173 to 0.3771 for video continuation, with trajectory error dropping from 0.3960 to 0.3830 and from 0.4956 to 0.4289. 
Although the GeoAnch-only variant achieves slightly higher I2V background PSNR and SSIM, the full model attains the best overall quality, camera control, and dynamic persistence in both tasks, confirming the complementary roles of GeoAnch and TemComp.

\noindent\textbf{Effect of Panoramic Sampling Density.}
We evaluate each mixed-interval checkpoint across $s\in\{1,2,4,10\}$ without retraining, varying the panoramic length over 81, 41, 21, and 9 frames (see Tables~\ref{tab:main-i2v} and~\ref{tab:main-continuation}). 
For CameraCtrl, $s=4$ achieves the lowest FID and rotation error in both tasks, whereas denser sampling benefits dynamic-persistence metrics. 
Conversely, UCPE benefits more from denser panoramic prediction, with $s=1$ yielding the lowest FID and rotation error. 
Meanwhile, $s=2$ maximizes R-ID in both tasks, increasing from 0.3694 at $s=4$ to 0.3839 for I2V and from 0.3849 to 0.4667 for continuation. 
Although trends are non-monotonic across metrics, even $s=10$ (9 frames) improves all 14 metrics over the baselines for both interfaces and tasks. 

\noindent\textbf{Analysis of Computational Cost.}
Table~\ref{tab:flops} reports per-step denoising FLOPs, runtime, and peak memory for an $81\times480\times832$ perspective video and an $(80/s+1)\times512\times1024$ panoramic video, including camera-control and cross-view modules. 
Sparse panoramic evolution achieves substantial gains with modest overhead. 
At $s=10$, the CameraCtrl and UCPE variants increase FLOPs by only 25.5\% and 27.2\%, runtime by 30.2\% and 29.2\%, and peak memory by only 1.0~GB and 0.7~GB over their baselines. 
At $s=4$, they reduce FLOPs by 50.5\% and 51.2\% and runtime by 43.5\% and 44.8\% compared with $s=1$, while adding only 1.1~GB and 0.8~GB in memory over the baselines. 
Furthermore, GeoAnch and TemComp each add $\sim$1~TFLOP (less than 0.7\% of the total cost) in $s=4$, maintaining efficiency.
Thus, OmniPersist offers a practical trade-off among visual fidelity, re-entry consistency, and computation.

\section{Conclusion}
\label{sec:conclusion}
In this paper, we address the critical challenge of dynamic object persistence in camera-controlled world models, where out-of-view moving entities frequently mutate or vanish upon viewpoint revisits. 
To resolve this limitation, we present \method{}, a single-stage framework that couples perspective video generation with $360^\circ$ panoramic evolution within a unified diffusion model, requiring no initial panoramic images or auxiliary models. 
By operating the panoramic stream at a lower frame rate, \method{} efficiently supports plausible evolution of out-of-view content. 
We seamlessly bridge the two streams across space and time via Geometric Anchoring (GeoAnch) for synchronous spatial correspondence and Temporal Compensation (TemComp) for cross-frame temporal aggregation. 
Extensive experiments show that \method{} achieves consistent performance gains across both latent-level (CameraCtrl) and attention-level (UCPE) camera conditioning in visual fidelity, camera controllability, background revisit consistency, and dynamic persistence.

\noindent\textbf{Limitations and Future Work.}
While \method{} sustains dynamic persistence during inference without requiring panoramic inputs, its training currently relies on panoramic video data. 
Nevertheless, with the increasing prevalence of omnidirectional capture hardware and realistic 3D simulation platforms, high-quality panoramic video assets are expected to expand rapidly. 
In future work, we plan to extend this framework toward real-time streaming and long-horizon video generation, and investigate its applications in robotic navigation.

\label{maintextend}

\clearpage




\renewcommand{\thesection}{\Alph{section}}
\renewcommand{\thetable}{\Alph{table}}
\renewcommand{\thefigure}{\Alph{figure}}
\renewcommand{\theequation}{\Alph{equation}}
\setcounter{section}{0}
\setcounter{figure}{0}
\setcounter{table}{0}
\setcounter{equation}{0}

\maketitlesupplementary


\noindent The content of the appendix involves: \vspace{0.7em}

\begin{itemize}[leftmargin=1.5em, label=--, itemsep=0.7em]
\item Sec.~\ref{sec_suppl:dataset} describes the dataset construction pipeline.
\item Sec.~\ref{sec_suppl:metrics} specifies evaluation metric implementations.
\item Sec.~\ref{sec_suppl:implementation} details the implementation of our framework.
\item Sec.~\ref{sec_suppl:results} provides additional ablation studies, quantitative evaluations, and qualitative results.
\end{itemize}

\section{Dataset Construction}
\label{sec_suppl:dataset}

\subsection{Overview of the Pipeline}

Our curation pipeline converts high-resolution raw panoramic footage into synchronized pairs of equirectangular projection (ERP) and perspective video clips. 
The panoramic stream is standardized by downsampling the footage to a resolution of $1024\times512$, with its camera-to-world poses calibrated using ViPE~\cite{huang2025vipe}. 

To simulate realistic camera viewing trajectories with intentional leave-and-revisit patterns, we ground camera motion on moving foreground subjects. 
Specifically, we employ SAM 3.1~\cite{carion2026sam} to track dynamic entities directly across the ERP frames. 
The initial camera orientation is anchored to a reliably tracked subject exhibiting sufficient pixel area near the panoramic center, establishing a stable visual prefix. 
If the subject is initially located off-center, we apply a horizontal rotation to the panorama to bring it to the central meridian. 
Guided by the motion paths of tracked subjects, we synthesize four categories of camera trajectories: object tracking, single revisit, double revisit, and full $\sim$360$^\circ$ camera rotations.

Perspective video clips are rendered from the panoramic footage along the generated viewing trajectories and scaled to a resolution of $832\times480$. 
To preserve visual fidelity and respect source-resolution constraints, the horizontal field of view (FoV) is capped at a maximum of $120^\circ$, while the minimum horizontal FoV is restricted to ensure that the natively sampled resolution is no less than half of $832\times480$ (limiting upscaling to a factor of two). 
The vertical FoV is determined by the $832:480$ aspect ratio. 

The absolute camera pose of each perspective frame is computed by compounding the calibrated panoramic pose with the simulated relative viewport transformation. 
Specifically, at discrete frame index $t$, the camera-to-world poses satisfy the geometric relationship:
\begin{equation}
 \mathbf{C}_t^{\mathrm{x}} = \mathbf{C}_t^{\mathrm{p}} \mathbf{V}_t,
 \qquad \mathbf{C}_t^{\mathrm{x}}, \mathbf{C}_t^{\mathrm{p}}, \mathbf{V}_t \in \mathbb{R}^{4\times4},
 \label{eq:pose-composition}
\end{equation}
where $\mathbf{C}_t^{\mathrm{x}}$ denotes the absolute camera-to-world pose of the perspective frame, $\mathbf{C}_t^{\mathrm{p}}$ represents the calibrated pose of the panoramic frame, and $\mathbf{V}_t$ is the homogeneous pure-rotation matrix governing the relative viewport orientation. 
Both pose sequences are defined in a shared world coordinate system anchored to the coordinate frame of the initial perspective camera, such that $\mathbf{V}_0 = \mathbf{I}$.

Because $\mathbf{V}_t$ contains solely rotational components, it steers the perspective viewing direction while strictly preserving the optical center trajectory of the source panorama. 
Consequently, physical translations of the panoramic capture device are faithfully preserved in the perspective clips, whereas the rendering process introduces no artificial translational offsets or synthesized motion parallax, guaranteeing physically grounded perspective observations.

\subsection{Camera Trajectory Synthesis}
\label{sec_suppl:trajectories}

We synthesize perspective camera trajectories across four semantic categories, defined by ten parameterized programs $\mathcal{T}$ as summarized in Table~\ref{tab:supp-trajectories}. 
These categories include object tracking ($\mathcal{T}_{\text{trk}}$) as a control baseline with continuous tracking, single revisit ($\mathcal{T}_{\text{sr}, 1\dots4}$) where the camera turns away from the subject once and returns, double revisit ($\mathcal{T}_{\text{dr}, 1\dots4}$) with two leave-and-return cycles, and full rotation ($\mathcal{T}_{\text{rot}}$) performing an approximately $360^\circ$ panoramic turn.

\begin{table}[t]
\centering
\caption{\textbf{Camera trajectory programs.} The ten parameterized programs are categorized into the four semantic trajectory types described in the main paper.}
\label{tab:supp-trajectories}
\small
\setlength{\tabcolsep}{3pt}
\begin{tabular}{@{}p{.32\columnwidth}lp{.52\columnwidth}@{}}
\toprule
Trajectory Type & Notation & Trajectory Pattern \\
\midrule
Object tracking & $\mathcal{T}_{\text{trk}}$ & Continuous subject tracking \\
\midrule
Single revisit & $\mathcal{T}_{\text{sr}, 1}$ & Horizontal pan and return \\
 & $\mathcal{T}_{\text{sr}, 2}$ & Asymmetric turn with pitch tilt \\
 & $\mathcal{T}_{\text{sr}, 3}$ & Vertical detour \\
 & $\mathcal{T}_{\text{sr}, 4}$ & Return overshoot \\
\midrule
Double revisit & $\mathcal{T}_{\text{dr}, 1}$ & Double horizontal pan and return \\
 & $\mathcal{T}_{\text{dr}, 2}$ & Double asymmetric turn with tilt \\
 & $\mathcal{T}_{\text{dr}, 3}$ & Double vertical detour \\
 & $\mathcal{T}_{\text{dr}, 4}$ & Double return overshoot \\
\midrule
Full $\sim$360$^\circ$ rotation & $\mathcal{T}_{\text{rot}}$ & $\sim$360$^\circ$ rotation \\
\bottomrule
\end{tabular}
\end{table}

To generate smooth and realistic camera motion, each trajectory interpolates yaw, pitch, and roll keyframes using piecewise quintic Hermite splines, ensuring continuous angular velocity and acceleration. 
Crucially, the reference camera orientation dynamically follows the moving subject rather than remaining static in world coordinates. 
Camera offsets are added on top of this moving reference orientation. 
Consequently, when the camera returns, it points to the current position of the subject instead of resetting to the initial orientation at the start of the video.

Single-revisit trajectories share an out-and-back structure with four motion patterns. 
Horizontal pan-and-return trajectories ($\mathcal{T}_{\text{sr}, 1}$) place the maximum yaw turn at the midpoint, producing symmetric departure and return phases.
Asymmetric turns ($\mathcal{T}_{\text{sr}, 2}$) shift the peak turn away from the temporal midpoint and add a small pitch tilt, producing unequal departure and return durations alongside vertical angle changes. 
Vertical detours ($\mathcal{T}_{\text{sr}, 3}$) combine the horizontal sweep with alternating positive and negative pitch waypoints, guiding the camera through an out-of-plane detour rather than a flat path. 
Return overshoots ($\mathcal{T}_{\text{sr}, 4}$) swing the camera past the reference heading to the opposite side before correcting back to the target direction.

Double revisits ($\mathcal{T}_{\text{dr}, 1\dots4}$) repeat this process twice, separating the two turns with a brief pause. 
To avoid visual distortion during fast turns, we limit the maximum angular speed to $90^\circ$/s for single revisits ($\mathcal{T}_{\text{sr}}$), $105^\circ$/s for double revisits ($\mathcal{T}_{\text{dr}}, \mathcal{T}_{\text{trk}}$), and $120^\circ$/s for full rotation ($\mathcal{T}_{\text{rot}}$). 
If a planned trajectory exceeds these speed limits, pauses and transition times are adjusted.
If the speed is still too high, the rotation angle is scaled down accordingly.

\subsection{Dataset Statistics}
\label{sec_suppl:dataset_statistics}

\noindent\textbf{Dataset Scale.}
The curated dataset comprises 125K training clips and 300 evaluation clips, with strictly disjoint source videos between the two splits. 
Multiple perspective clips are rendered from individual panoramic source videos under different camera trajectories. 
Across the dataset, the ten parameterized trajectory programs in Table~\ref{tab:supp-trajectories} are sampled uniformly, ensuring balanced representation.

\begin{figure}[t]
    \centering
    \begin{subfigure}[b]{.47\columnwidth}
        \centering
        \includegraphics[width=\linewidth,height=.85\linewidth,keepaspectratio]{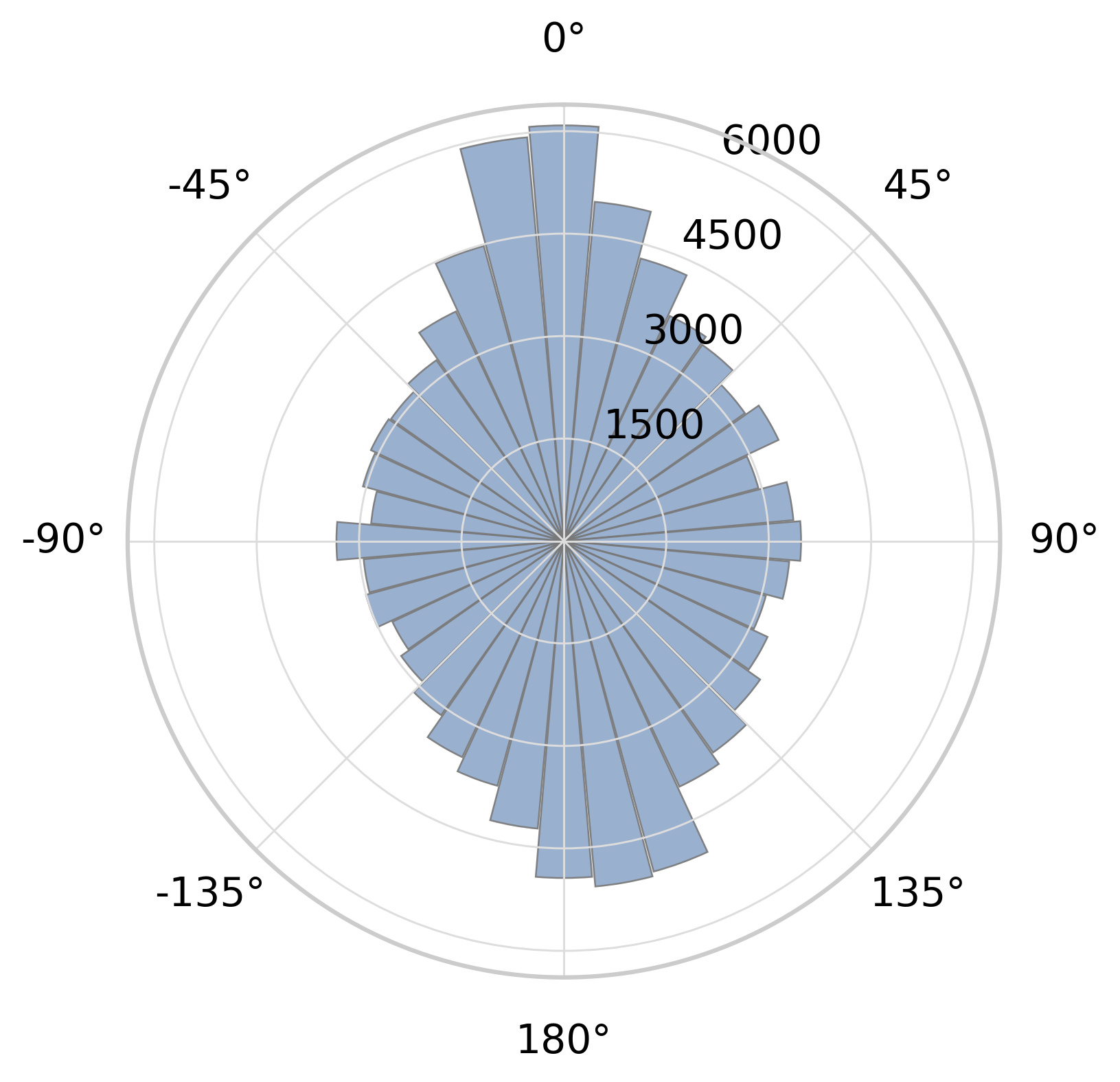}
        \caption{Motion direction}
        \label{fig:supp-camera-motion-direction}
    \end{subfigure}\hfill
    \begin{subfigure}[b]{.47\columnwidth}
        \centering
        \includegraphics[width=\linewidth,height=.85\linewidth,keepaspectratio]{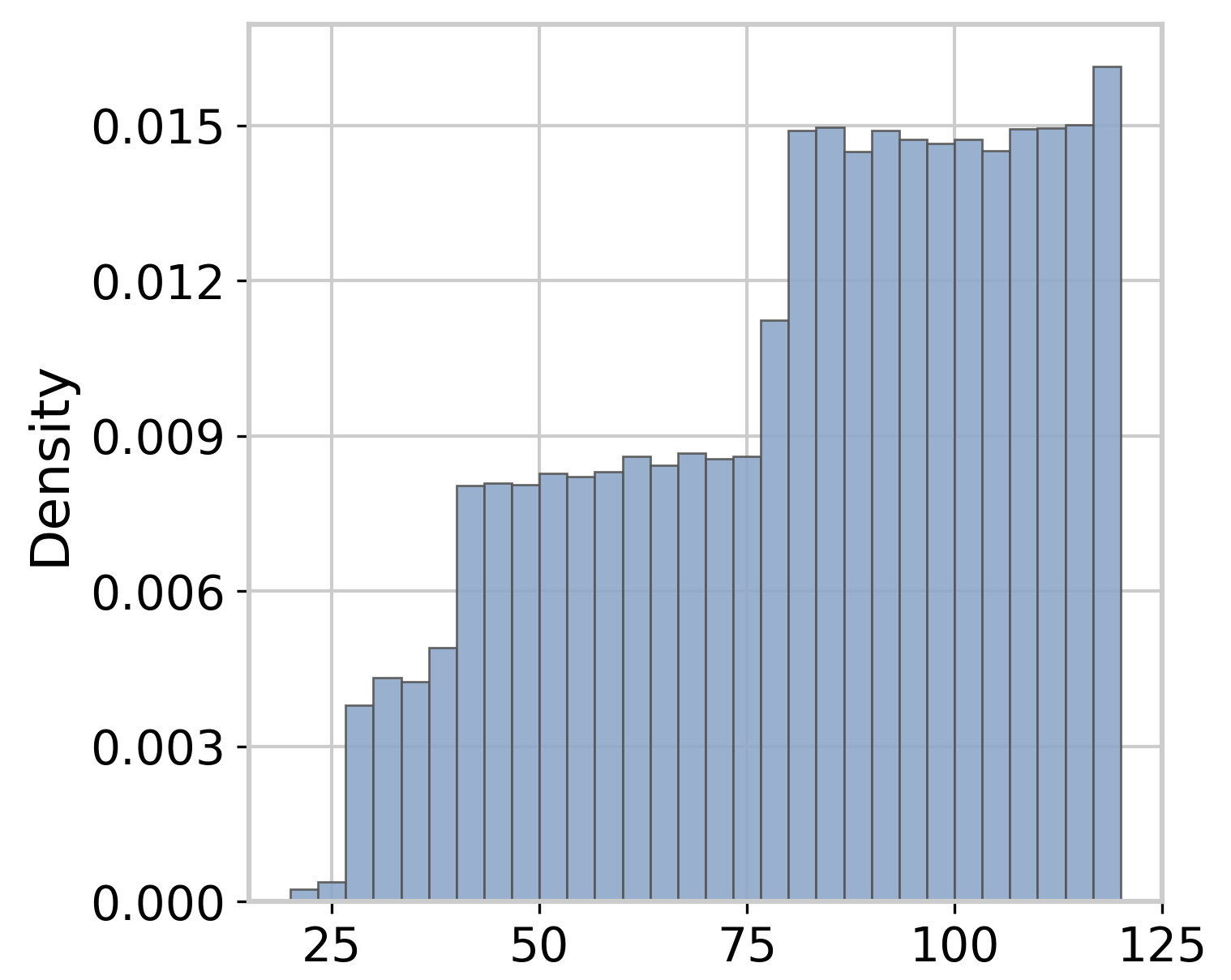}
        \caption{Horizontal FoV}
        \label{fig:supp-fov-horizontal}
    \end{subfigure}
    \caption{\textbf{Camera statistics of the training set.} (a) Polar distribution of camera motion directions, where radial bar lengths denote sample counts across $360^\circ$ heading angles. (b) Probability density of the horizontal perspective FoV in degrees. The dataset exhibits broad and balanced coverage across both motion directions and viewing angles.}
    \label{fig:supp-camera-statistics}
\end{figure}

\noindent\textbf{Camera Motion Direction Distribution.}
We analyze the distribution of camera motion directions across the training set to assess the diversity of generated camera paths. 
For each clip, the camera motion direction is computed from the net horizontal rotation between the first and last frames. 
As illustrated in Figure~\ref{fig:supp-camera-statistics}(a), the motion directions span the complete $360^\circ$ range with broad coverage. 
Because revisit trajectories where the camera returns toward the subject constitute a major proportion of the dataset, the distribution naturally exhibits higher density around $0^\circ$ and $180^\circ$, while maintaining substantial representation across all lateral angles.

\noindent\textbf{Field of View Distribution.}
Figure~\ref{fig:supp-camera-statistics}(b) illustrates the horizontal field of view (FoV) distribution across the training set. 
Because source panoramic videos vary in resolution and perspective sampling strictly enforces an upscaling limit, the resulting FoV distribution exhibits a tiered density rather than a strictly uniform spread. 
Nevertheless, the horizontal FoV spans broadly from approximately $25^\circ$ to the preset upper limit of $120^\circ$, providing rich variations in focal lengths and visual perspective. 
The distribution maintains substantial coverage across both standard viewing angles and wide-angle configurations. 
This broad FoV coverage prevents models from overfitting to a fixed focal length and encourages generalization across diverse perspective geometries.

\section{Evaluation Metrics}
\label{sec_suppl:metrics}

During evaluation, the known prefix $\mathbf{X}_{0:K-1}$ provides conditioning context but is excluded from quantitative metric computation, where evaluated frame indices span $1$--$80$ for image-to-video (I2V) and $21$--$80$ for video continuation. 

\subsection{Overall Quality}
We evaluate overall visual quality using FID~\cite{heusel2017fid}, FVD~\cite{unterthiner2018fvd}, and center-cropped FVD (FVD$_c$). 
Specifically, FID is computed on 2048-dimensional feature representations from Inception-v3~\cite{szegedy2016rethinking}, while FVD is calculated on 400-dimensional classification logits from a pretrained I3D network~\cite{carreira2017quo}. 
Standard FVD resizes video frames directly to $224\times224$, whereas FVD$_c$ preserves the aspect ratio by scaling the shorter spatial dimension to 224 pixels and applying a central crop. 
Both variants compute the Fr\'echet distance between Gaussian distributions fitted to feature representations of generated and ground-truth videos.

\subsection{Camera Controllability}
We evaluate camera controllability using rotation error (Rot), translation error (Trans)~\cite{he2025cameractrl}, CamMC~\cite{wang2024motionctrl}, and absolute trajectory error (ATE). 
ViPE~\cite{huang2025vipe} independently reconstructs the camera extrinsics of both generated and ground-truth (GT) videos using known camera intrinsics. 
All pose metrics operate on relative trajectories anchored at the initial frame. 
Let $\hat{\mathbf{R}}_t, \mathbf{R}_t \in \mathbb{R}^{3\times3}$ and $\hat{\mathbf{p}}_t, \mathbf{p}_t \in \mathbb{R}^3$ denote the anchored rotations and translations for the generated and ground-truth sequences, respectively. 

Following UCPE~\cite{zhang2026ucpe}, translations are independently normalized by their maximum trajectory radius prior to score computation:
\begin{equation}
\begin{aligned}
 \hat{\mathbf{p}}^{\star}_t &= \frac{\hat{\mathbf{p}}_t}{\max_{0 \le r < T}\|\hat{\mathbf{p}}_r\|_2 + 10^{-9}}, \\
 \mathbf{p}^{\star}_t &= \frac{\mathbf{p}_t}{\max_{0 \le r < T}\|\mathbf{p}_r\|_2 + 10^{-9}}.
\end{aligned}
\label{eq:trans-norm}
\end{equation}
For the $N$ evaluated suffix frames, the quantitative errors are given by
\begin{align}
 E_{\mathrm{Rot}} &= \frac{180}{\pi N}\sum_t\arccos\!\left(
 \frac{\operatorname{tr}(\hat{\mathbf{R}}_t^\top\mathbf{R}_t)-1}{2}\right), \\
 E_{\mathrm{Trans}} &= \frac{1}{N}\sum_t\|\hat{\mathbf{p}}^{\star}_t-\mathbf{p}^{\star}_t\|_2, \\
 E_{\mathrm{ATE}} &= \sqrt{\frac{1}{N}\sum_t\|\hat{\mathbf{p}}^{\star}_t-\mathbf{p}^{\star}_t\|_2^2}, \\
 E_{\mathrm{CamMC}} &= \frac{1}{N}\sum_t\|\operatorname{vec}(\hat{\mathbf{C}}^{\star}_t)-\operatorname{vec}(\mathbf{C}^{\star}_t)\|_2,
\end{align}
where the argument of the arccosine is numerically clamped to $[-1, 1]$. 
Here, $\mathbf{C}_t^{\star} = [\mathbf{R}_t, \mathbf{p}_t^{\star}] \in \mathbb{R}^{3\times4}$ concatenates rotation and normalized translation (and analogously for $\hat{\mathbf{C}}_t^{\star}$), which $\operatorname{vec}(\cdot)$ flattens into a 12-dimensional vector.

\subsection{Background Revisit Fidelity}

Revisit timestamps are identified deterministically from the ground-truth camera trajectory, independent of generated video content. 
For each candidate frame $t \ge 32$, we identify its closest conditioning view $j_t^\star = \arg\min_{j<K} d(t,j)$, with $K=1$ for I2V and $K=21$ for video continuation, using ground-truth camera poses and the normalized pose distance
\begin{equation}
 d(t,j) = \frac{\angle(\mathbf{R}_t,\mathbf{R}_j)}{5^\circ} + \frac{\|\mathbf{p}_t-\mathbf{p}_j\|_2}{0.05D},
 \label{eq:revisit-dist}
\end{equation}
where $\angle(\cdot,\cdot)$ denotes the geodesic rotation angle, $\mathbf{p}_t$ represents the camera center, and $D=\max_{a,b}\|\mathbf{p}_a-\mathbf{p}_b\|_2$ is the maximum pairwise distance along the ground-truth trajectory. 
We rank candidate frames by $d(t,j_t^\star)$ and greedily select up to five frames, enforcing a minimum temporal separation of four frames between chosen timestamps.  
Frame selections are precomputed separately for the two conditioning lengths and shared across all evaluated methods.

At each selected timestamp, we evaluate PSNR, SSIM~\cite{wang2004ssim}, and VGG-based~\cite{simonyan2014very} LPIPS~\cite{zhang2018unreasonable} between generated frame and the corresponding ground-truth frame, restricting comparison to the intersection of their static-background masks. 
Frame pairs with less than 10\% valid background coverage are excluded from evaluation. 

\subsection{Dynamic Subject Persistence} 

We quantify dynamic persistence using overall subject identity (ID), re-entry identity (R-ID), visibility F1-score (Vis.), and normalized subject trajectory error (Traj.). 

\noindent\textbf{Instance Tracking and Event Annotation.}
We first track foreground objects across ground-truth (GT) videos using SAM~3.1~\cite{carion2026sam}. 
For each video, we select up to five salient instances present in the initial frame that successfully re-appear following camera revisits. 
To determine visibility transitions, we track the same instances across the ground-truth panoramic video and reproject each panoramic instance mask onto the perspective view using the relative camera transformation at that frame. 
When panoramic tracking is reliable but the reprojected mask area on the perspective image plane is zero, the timestamp is labeled as an out-of-view event. 
Conversely, a non-zero reprojected area signifies an in-view state, establishing continuous visibility event annotations throughout the sequence.

\noindent\textbf{Subject and Re-Entry Identity.}
In the corresponding generated video, independent tracking is initialized using identical first-frame instance prompts.
Overall subject identity (ID) measures the cosine similarity between foreground feature representations extracted via DINOv2~\cite{oquab2023dinov2} from the generated and the corresponding ground-truth instances. 
If the ground-truth mask is invalid at a given timestamp, we search for the nearest valid ground-truth frame within a temporal window of $\pm 10$ frames to maintain reliable feature comparisons. 
Re-entry identity (R-ID) specifically isolates the critical phase of subject recovery after an out-of-view period. 
The evaluation interval commences at the ground-truth re-entry timestamp and extends until the subject exits the field of view again. 
Within this interval, we collect up to the earliest eight valid ground-truth frames to evaluate re-entry identity, directly assessing whether the subject preserves its appearance immediately upon returning.

\noindent\textbf{Visibility F1-Score.}
In the generated video, an instance is considered visible when object tracking produces a non-empty mask with valid tracking quality. The instance is confirmed to be out of frame when the mask touches the image boundary before disappearing.
The visibility F1-score is defined as $2\mathrm{TP}/(2\mathrm{TP}+\mathrm{FP}+\mathrm{FN})$, evaluated between generated and GT visibility states across frame indices with deterministic GT states.

\noindent\textbf{Subject Trajectory Error.}
Because generated videos are primarily intended for viewing in 2D space, we employ the bounding-box center discrepancy between generated and GT objects to measure the accuracy of subject motion prediction in the main text.
The metric compares bounding-box centers after rescaling predicted coordinates to the GT image dimensions, defined as
\begin{equation}
 e_t^{\mathrm{2D}}=\begin{cases}
 \|\hat{\mathbf c}_t-\mathbf c_t\|_2/\sqrt{W^2+H^2},
 &\text{reliable prediction},\\
 1,&\text{otherwise},
 \end{cases}
\end{equation}
where $\hat{\mathbf c}_t$ and $\mathbf c_t$ denote the predicted and GT bounding-box centers at frame $t$, respectively, and $W$ and $H$ represent the width and height of the GT frame. Missing generated subjects and failed track matches receive a penalty of $1$, whereas frames with unavailable GT positions are excluded from evaluation.

\subsection{Subject 3D Displacement Error}
\label{suppl_sec:3d_error}

The subject trajectory error reported in the main paper may be affected by the accuracy of camera pose predictions. Therefore, we additionally report the relative 3D displacement error of each object to evaluate dynamic subject persistence. Quantitative results are presented in Sec.~\ref{suppl_sec:3d_eval}, while this section details the metric formulation.

We report \textbf{3D Displacement Error (3D-DE$\downarrow$)} and \textbf{3D Displacement Coverage (3D-DC$\uparrow$)} as two complementary metrics based on independently reconstructed geometry for ground-truth and generated videos. We use UniDepth V2~\cite{piccinelli2025unidepthv2} to estimate optical-axis depth while reusing existing camera poses from ViPE.
For each instance, we back-project the pixels with valid positive depth within its mask into the initial-camera coordinate system of the video. Their coordinate-wise median defines a robust surface-position proxy $\mathbf{m}_t$. Let
\begin{equation}
\boldsymbol{\delta}_t=\mathbf{m}_t-\mathbf{m}_0, \quad
\hat{\boldsymbol{\delta}}_t=
\hat{\mathbf{m}}_t-\hat{\mathbf{m}}_0,  \quad
L=(\mathbf{m}_0)_z>0,
\end{equation}
where unaccented and hatted quantities denote ground truth and prediction, respectively, and $L$ represents the depth of the GT object at frame zero.

Let $\mathcal{T}$ contain the evaluated suffix frames with confirmed ground-truth subject presence and reliable ground-truth tracking and geometry at frames zero and $t$. For $t\in\mathcal{T}$, let $v_t=1$ if the prediction also has reliable tracking and valid 3D positions at both timestamps, and $v_t=0$ otherwise. We define
\begin{equation}
e_t^{\mathrm{3D}}=
\begin{cases}
\displaystyle
\min\!\left(
\frac{\|\hat{\boldsymbol{\delta}}_t-
\boldsymbol{\delta}_t\|_2}{L},\,1
\right), & v_t=1,\\[4pt]
1, & v_t=0,
\end{cases}
\end{equation}
and the per-subject metrics are
\begin{equation}
\mathrm{3D\text{-}DE}
=\frac{1}{|\mathcal{T}|}\sum_{t\in\mathcal{T}}e_t^{\mathrm{3D}},
\qquad
\mathrm{3D\text{-}DC}
=\frac{1}{|\mathcal{T}|}\sum_{t\in\mathcal{T}}v_t.
\end{equation}
Normalizing by the initial depth $L$ of the ground-truth subject rather than its motion amplitude avoids instability for nearly stationary subjects. Coverage captures the availability of valid 3D displacement estimates. To some extent, higher values reflect more consistent subject presence and more reliable 3D geometry in the generated video. These metrics therefore provide evidence complementary to the 2D trajectory and visibility scores.

\section{Implementation Details}
\label{sec_suppl:implementation}
This section details temporal indexing and latitude weighting in loss calculation.

\subsection{Temporal Indexing}
Under the $4\times$ temporal compression of the shared VAE~\cite{kingma2014auto}, a $T$-frame perspective video and its $J$-frame panoramic counterpart, sampled at interval $s$ with $J=(T-1)/s+1$, yield latent lengths
\begin{equation}
 F_{\mathrm x}=\frac{T-1}{4}+1,\qquad
 F_{\mathrm p}=\frac{J-1}{4}+1=\frac{T-1}{4s}+1.
\end{equation}
Perspective latent index $i$ and panoramic latent index $j$ correspond to original-video reference times $4i$ and $4sj$, respectively. These serve as reference coordinates for temporally encoded features.

\begin{table}[t!]
\centering
\caption{\textbf{Panoramic temporal RoPE for an 81-frame perspective video.} The perspective stream consistently has 21 latent frames. Panoramic RoPE positions are assigned to align with the perspective latent timeline.}
\label{tab:supp-time}
\small
\begin{tabular}{@{}cccc@{}}
\toprule
$s$ & Pano RGB $J$ & Pano latent $F_{\mathrm p}$ & Pano RoPE positions \\
\midrule
1 & 81 & 21 & $0,1,\ldots,20$ \\
2 & 41 & 11 & $0,2,\ldots,20$ \\
4 & 21 & 6 & $0,4,8,12,16,20$ \\
10 & 9 & 3 & $0,10,20$ \\
\bottomrule
\end{tabular}
\end{table}

\begin{table*}[t!]
\centering
\caption{\textbf{Comparison with a single static panoramic state.} The experiments are conducted based on CameraCtrl~\cite{he2025cameractrl}.}
\label{tab:supp-pano-evolution}
\scriptsize
\setlength{\tabcolsep}{2pt}
\resizebox{\textwidth}{!}{%
\begin{tabular}{@{}lcccccccccccccc@{}}
\toprule
\multirow{2}{*}{Configuration} & \multicolumn{3}{c}{Overall Quality} & \multicolumn{4}{c}{Camera Controllability} & \multicolumn{3}{c}{BG Revisit Fidelity} & \multicolumn{4}{c}{Dynamic Subject Persistence} \\
\cmidrule(lr){2-4}\cmidrule(lr){5-8}\cmidrule(lr){9-11}\cmidrule(lr){12-15}
 & FID\dn & FVD\dn & FVD$_c$\dn & Rot\dn & Trans\dn & CamMC\dn & ATE\dn & PSNR\up & SSIM\up & LPIPS\dn & ID\up & R-ID\up & Vis.\up & Traj.\dn \\
\midrule
\multicolumn{15}{l}{\cellcolor{gray!15}\textit{Image-to-video ($K=1$)}} \\
w/o Pano & 23.60 & 288.5 & 295.1 & 13.086 & 0.531 & 0.677 & 0.594 & 12.29 & 0.4415 & 0.618 & 0.5834 & 0.2325 & 0.6519 & 0.4753 \\
Single panoramic state & 31.19 & 295.5 & 309.0 & 12.628 & 0.508 & 0.649 & 0.564 & 12.68 & 0.4451 & 0.628 & 0.5816 & 0.2575 & 0.6526 & 0.4870 \\
\method{} ($s=10$) & 22.91 & 266.6 & 276.4 & 5.695 & 0.418 & 0.478 & 0.465 & 13.37 & 0.4731 & 0.580 & 0.6429 & 0.2840 & 0.7333 & 0.3783 \\
\method{} ($s=4$) & 22.18 & 262.3 & 252.6 & 5.556 & 0.399 & 0.458 & 0.446 & 13.58 & 0.4754 & 0.573 & 0.6395 & 0.3355 & 0.7290 & 0.3830 \\
\midrule
\multicolumn{15}{l}{\cellcolor{gray!15}\textit{Video continuation ($K=21$)}} \\
w/o Pano & 22.61 & 227.5 & 223.5 & 11.704 & 0.451 & 0.597 & 0.489 & 13.43 & 0.4701 & 0.579 & 0.4946 & 0.2494 & 0.6053 & 0.5161 \\
Single panoramic state & 34.58 & 274.7 & 280.5 & 10.291 & 0.468 & 0.590 & 0.507 & 13.58 & 0.4635 & 0.600 & 0.4773 & 0.3293 & 0.5971 & 0.5411 \\
\method{} ($s=10$) & 21.69 & 201.0 & 200.6 & 4.661 & 0.401 & 0.455 & 0.435 & 15.20 & 0.5221 & 0.518 & 0.5559 & 0.3492 & 0.6835 & 0.4287 \\
\method{} ($s=4$) & 21.59 & 194.9 & 201.9 & 4.345 & 0.401 & 0.450 & 0.433 & 15.22 & 0.5238 & 0.515 & 0.5534 & 0.3771 & 0.6678 & 0.4289 \\
\bottomrule
\end{tabular}%
}
\end{table*}

\noindent\textbf{Temporal RoPE Alignment.}
Temporal RoPE~\cite{su2024roformer} coordinates $i$ and $sj$ for perspective and panoramic latent frames are constructed separately before concatenating the two token sequences. Meanwhile, spatial RoPE retains the two-dimensional token grid of each view. Table~\ref{tab:supp-time} summarizes the resulting panoramic RoPE positions across sampling intervals for $T=81$.

\noindent\textbf{Camera Alignment.}
In Eq.~(1) of the main paper, the camera intrinsics $\mathbf A_i$ and rotations $\mathbf R_i^{\mathrm x}$ attached to perspective latent frame $i$ are sampled from perspective RGB frame $4i$, whereas the panoramic rotation $\mathbf R_j^{\mathrm p}$ attached to panoramic latent frame $j$ is sampled from panoramic RGB frame $4j$. At synchronized indices $i=sj$, both views refer to original time $4sj$ and share a camera center. 

\subsection{Latitude Weighting}

Let $H_{\mathrm p}$ be the panoramic VAE latent height, and let $y\in\{0,\ldots,H_{\mathrm p}-1\}$ denote the row index. Following Argus~\cite{luo2025beyond}, we construct the latitude-dependent row weights as
\begin{align}
 q_y&=\frac{|y-(H_{\mathrm p}/2-1/2)|}{H_{\mathrm p}/2},\\
 a_y&=\sqrt{1-q_y^2}+\alpha,\qquad \alpha=0.25,\\
 w_y&=\frac{a_y}{\sum_{r=0}^{H_{\mathrm p}-1}a_r / H_{\mathrm p}},
\end{align}
where $q_y$ represents the normalized vertical distance from the equator, $a_y$ is the unnormalized row weight with a constant offset $\alpha=0.25$ to prevent zero weights near the poles, and $w_y$ denotes the normalized row weight. These weights $w_y$ emphasize equatorial rows while retaining positive weights near the poles.

In Eq.~(4) of the main paper, the latitude-dependent weight $\mathbf W_{\mathrm p}$ inside the squared norm is defined row-wise as $(\mathbf W_{\mathrm p})_y=\sqrt{w_y}$. Thus, the panoramic loss $\mathcal L_{\mathrm p}$ can also be expressed as
\begin{equation}
 \mathcal L_{\mathrm p}=\operatorname{mean}\!\left[
 w_y(\hat{\mathbf u}^{\mathrm p}-\mathbf u^{\mathrm p})^2\right],
\end{equation}
where $\hat{\mathbf u}^{\mathrm p}$ and $\mathbf u^{\mathrm p}$ denote the predicted and target panoramic velocity fields, respectively.

\begin{table}[t!]
\begingroup
\centering
\captionsetup{position=top,skip=10pt}
\caption{\textbf{3D displacement evaluation for image-to-video (I2V).} Rows correspond to main-paper Table~1. 3D-DE measures 3D displacement error, and 3D-DC reports 3D tracking coverage.}
\vspace{-1mm}
\label{tab:supp-dynamic-i2v}
\fontsize{8.1}{9.5}\selectfont
\setlength{\tabcolsep}{2pt}
\begin{tabular*}{.86\columnwidth}{@{\extracolsep{\fill}}p{.54\columnwidth}c@{\hspace{10pt}}c@{}}
\toprule
Configuration & 3D-DE\dn & 3D-DC\up \\
\midrule
HyDRA (1.3B)~\cite{chen2026hybrid} & \secondcell{0.6150} & \secondcell{0.6153} \\
Matrix-World 3.0 (5B)~\cite{wang2026matrix} & 0.6929 & 0.4682 \\
DreamX-World (5B)~\cite{team2026dreamx} & 0.6702 & 0.5254 \\
SANA-WM (2.6B+17B)~\cite{zhu2026sana} & \thirdcell{0.6380} & \thirdcell{0.5562} \\
LiveWorld (14B)~\cite{duan2026liveworld} & 0.6755 & 0.5425 \\
LingBot-World 2.0 (28B)~\cite{gao2026infinite} & \firstcell{0.5490} & \firstcell{0.6913} \\
\midrule
CameraCtrl\textsuperscript{\dag}~\cite{he2025cameractrl} & 0.6380 & 0.5792 \\
CameraCtrl+OmniPersist ($s=1$) & \firstcell{0.5313} & \secondcell{0.6740} \\
CameraCtrl+OmniPersist ($s=2$) & \thirdcell{0.5359} & \firstcell{0.6785} \\
CameraCtrl+OmniPersist ($s=4$) & 0.5447 & \thirdcell{0.6735} \\
CameraCtrl+OmniPersist ($s=10$) & \secondcell{0.5358} & \secondcell{0.6740} \\
\midrule
UCPE\textsuperscript{\dag}~\cite{zhang2026ucpe} & 0.6094 & 0.5748 \\
UCPE+OmniPersist ($s=1$) & \secondcell{0.5326} & \thirdcell{0.6672} \\
UCPE+OmniPersist ($s=2$) & \thirdcell{0.5433} & \secondcell{0.6795} \\
UCPE+OmniPersist ($s=4$) & \firstcell{0.5304} & \firstcell{0.6863} \\
UCPE+OmniPersist ($s=10$) & 0.5565 & 0.6572 \\
\bottomrule
\end{tabular*}

\par\endgroup

\vspace{15pt}

\begingroup
\centering
\captionsetup{position=top,skip=10pt}
\caption{\textbf{3D displacement evaluation for video continuation.} Rows correspond to main-paper Table~2. 3D-DE measures 3D displacement error, and 3D-DC reports 3D tracking coverage.}
\vspace{-1mm}
\label{tab:supp-dynamic-continuation}
\fontsize{8.1}{9.5}\selectfont
\setlength{\tabcolsep}{2pt}
\begin{tabular*}{.86\columnwidth}{@{\extracolsep{\fill}}p{.54\columnwidth}c@{\hspace{10pt}}c@{}}
\toprule
Configuration & 3D-DE\dn & 3D-DC\up \\
\midrule
HyDRA (1.3B)~\cite{chen2026hybrid} & 0.8949 & \thirdcell{0.2622} \\
Matrix-World 3.0 (5B)~\cite{wang2026matrix} & \thirdcell{0.8412} & 0.2570 \\
SANA-WM (2.6B+17B)~\cite{zhu2026sana} & \secondcell{0.7979} & \secondcell{0.3800} \\
LiveWorld (14B)~\cite{duan2026liveworld} & \firstcell{0.7560} & \firstcell{0.3806} \\
\midrule
CameraCtrl\textsuperscript{\dag}~\cite{he2025cameractrl} & 0.6847 & 0.4965 \\
CameraCtrl+OmniPersist ($s=1$) & \firstcell{0.5775} & \firstcell{0.6077} \\
CameraCtrl+OmniPersist ($s=2$) & \secondcell{0.5984} & \secondcell{0.5964} \\
CameraCtrl+OmniPersist ($s=4$) & \thirdcell{0.6031} & \thirdcell{0.5853} \\
CameraCtrl+OmniPersist ($s=10$) & 0.6039 & 0.5847 \\
\midrule
UCPE\textsuperscript{\dag}~\cite{zhang2026ucpe} & 0.6795 & 0.4741 \\
UCPE+OmniPersist ($s=1$) & \secondcell{0.5752} & \secondcell{0.5916} \\
UCPE+OmniPersist ($s=2$) & \firstcell{0.5677} & \firstcell{0.6006} \\
UCPE+OmniPersist ($s=4$) & \thirdcell{0.5853} & \thirdcell{0.5889} \\
UCPE+OmniPersist ($s=10$) & 0.6080 & 0.5565 \\
\bottomrule
\end{tabular*}

\par\endgroup

\vspace{15pt}

\begingroup
\centering
\captionsetup{position=top,skip=10pt}
\caption{\textbf{3D displacement evaluation for component ablation.} Rows correspond to main-paper Table~3. 3D-DE measures 3D displacement error, and 3D-DC reports 3D tracking coverage.}
\vspace{-1mm}
\label{tab:supp-dynamic-components}
\fontsize{8.1}{9.5}\selectfont
\setlength{\tabcolsep}{2pt}
\begin{tabular*}{.86\columnwidth}{@{\extracolsep{\fill}}p{.54\columnwidth}c@{\hspace{10pt}}c@{}}
\toprule
Configuration & 3D-DE\dn & 3D-DC\up \\
\midrule
\multicolumn{3}{l}{\cellcolor{gray!15}\textit{Image-to-video ($K=1$)}} \\
w/o Pano & 0.6380 & 0.5792 \\
w/ Pano & \thirdcell{0.6101} & \thirdcell{0.5952} \\
w/ Pano \& GeoAnch & \secondcell{0.5548} & \secondcell{0.6644} \\
w/ Pano \& GeoAnch \& TemComp & \firstcell{0.5447} & \firstcell{0.6735} \\
\midrule
\multicolumn{3}{l}{\cellcolor{gray!15}\textit{Video continuation ($K=21$)}} \\
w/o Pano & \thirdcell{0.6847} & \thirdcell{0.4965} \\
w/ Pano & 0.6876 & 0.4858 \\
w/ Pano \& GeoAnch & \secondcell{0.6462} & \secondcell{0.5404} \\
w/ Pano \& GeoAnch \& TemComp & \firstcell{0.6031} & \firstcell{0.5853} \\
\bottomrule
\end{tabular*}

\par\endgroup

\end{table}

\section{Additional Results}
\label{sec_suppl:results}

\subsection{Comparison with Static Panoramic State}
We examine whether a single initial panoramic state can replace panoramic evolution. Table~\ref{tab:supp-pano-evolution} compares four CameraCtrl-based configurations on both tasks, namely the perspective-only baseline, joint prediction with a single panoramic state at original time $t=0$, and \method{} at $s=10$ and $s=4$. The single-state variant predicts only one panoramic frame throughout denoising. Although this configuration improves re-entry identity over the baseline in both tasks, it worsens FID, FVD, and subject trajectory error. Initial panoramic support alone therefore provides inconsistent benefits across generation quality and dynamic persistence.
In contrast, evolving panoramic states at both $s=10$ and $s=4$ outperform the single-state configuration across all 14 metrics in both tasks.

\subsection{3D Displacement Evaluation}
\label{suppl_sec:3d_eval}

Tables~\ref{tab:supp-dynamic-i2v}--\ref{tab:supp-dynamic-components} extend main-paper Tables~1--3 with 3D displacement evaluation. We report 3D displacement error (3D-DE, lower is better) and 3D tracking coverage (3D-DC, higher is better), as defined in Sec.~\ref{suppl_sec:3d_error}.

At the primary setting $s=4$, \method{} lowers 3D-DE and raises 3D-DC relative to both camera-control baselines in both tasks, aligning with the visibility and 2D trajectory gains in the main paper. Specifically, 3D-DE in I2V drops from 0.6380 to 0.5447 for CameraCtrl and from 0.6094 to 0.5304 for UCPE, while 3D-DC in video continuation under UCPE rises from 0.4741 to 0.5889. Both $s=4$ variants also surpass all released baselines on both metrics in video continuation and achieve lower error than LingBot-World 2.0 in I2V, though with slightly lower coverage. These 3D results support the persistence gains in the main comparison.
Gains over the perspective-only baselines persist across all four sampling intervals in both tasks, with even $s=10$ achieving lower 3D-DE and higher 3D-DC, confirming that sparse panoramic evolution remains effective.

The component ablation in Table~\ref{tab:supp-dynamic-components} aligns with the main-paper ablation. Adding a panoramic stream alone yields limited I2V gains and slightly degrades both 3D metrics in video continuation. Incorporating GeoAnch and TemComp progressively improves both measures across tasks, with TemComp reducing 3D-DE in video continuation from 0.6462 to 0.6031 and raising 3D-DC from 0.5404 to 0.5853, confirming the value of both modules.

\subsection{Additional Qualitative Results}

\begin{figure*}[t!]
\centering
\captionsetup[subfigure]{font=promptoverview,skip=4pt}
\begin{subfigure}[b]{\textwidth}
\centering
\includegraphics[width=\textwidth]{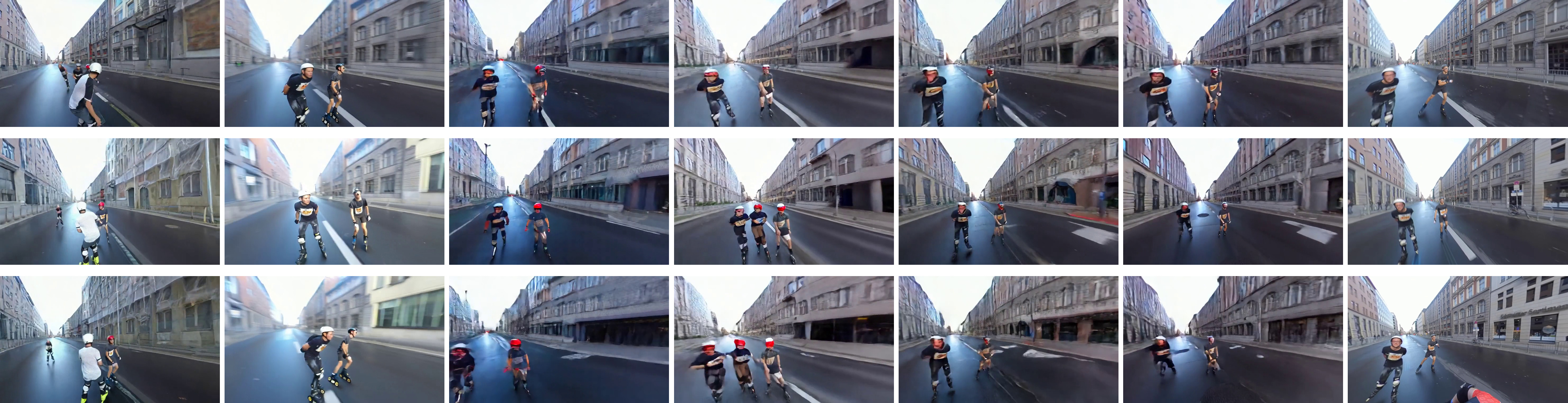}\par
\vspace{0pt}
{
\fontsize{7}{8}\selectfont
\setlength{\tabcolsep}{0pt}
\begin{tabular}{@{}*{7}{>{\centering\arraybackslash}p{\dimexpr\textwidth/7\relax}}@{}}
LiveWorld & LingBot-World 2.0 & CameraCtrl & UCPE & CameraCtrl+OmniPersist & UCPE+OmniPersist & Ground Truth \\
\end{tabular}
}
\vspace{-3mm}
\caption{Prompt overview: \textit{A video shows the skater in the white helmet and the skater in the grey helmet are positioned side-by-side, moving forward along the road.}}
\label{fig:supp-revisits-I2V-1}
\end{subfigure}\par
\vspace{3mm}
\begin{subfigure}[b]{\textwidth}
\centering
\includegraphics[width=\textwidth]{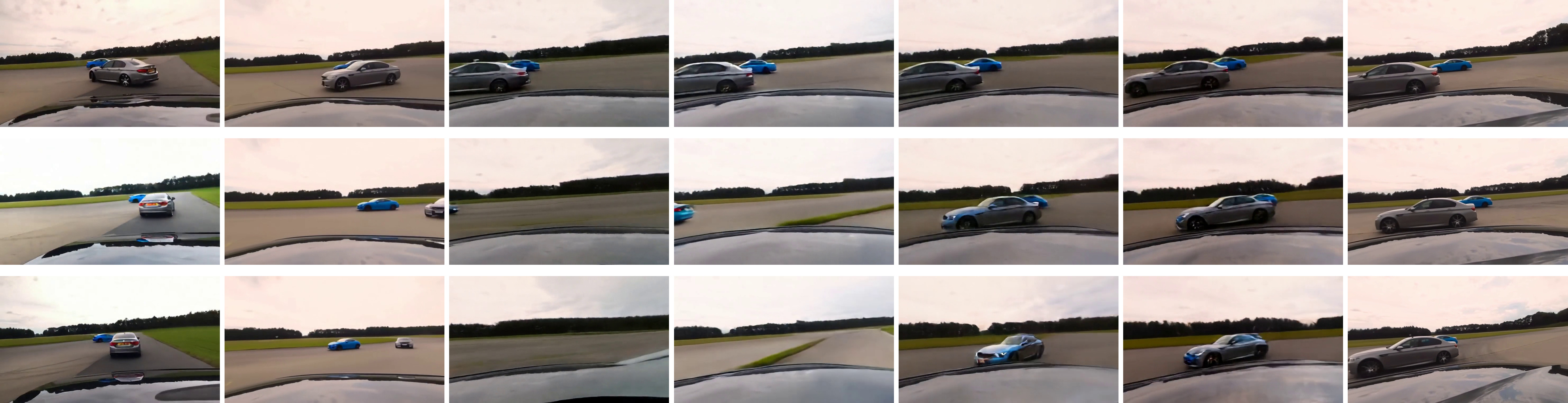}\par
\vspace{0pt}
{
\fontsize{7}{8}\selectfont
\setlength{\tabcolsep}{0pt}
\begin{tabular}{@{}*{7}{>{\centering\arraybackslash}p{\dimexpr\textwidth/7\relax}}@{}}
LiveWorld & LingBot-World 2.0 & CameraCtrl & UCPE & CameraCtrl+OmniPersist & UCPE+OmniPersist & Ground Truth \\
\end{tabular}
}
\vspace{-3mm}
\caption{Prompt overview: \textit{A video shows a silver sedan and a blue coupe driving on a paved track. The vehicles maintain a consistent distance from one another as they travel along the straight section of the track.}}
\label{fig:supp-revisits-I2V-2}
\end{subfigure}\par
\vspace{1mm}
\caption{\textbf{Generation results on I2V across camera revisits.} From top to bottom in each example, we show a frame before the camera turns away and two revisit frames after it returns to illustrate identity preservation and continued motion. Frames in each row share the same frame index across all methods and ground truth.}
\label{fig:supp-revisits-I2V}
\end{figure*}

\begin{figure*}[t!]
\centering
\captionsetup[subfigure]{font=promptoverview,skip=4pt}
\begin{subfigure}[b]{\textwidth}
\centering
\includegraphics[width=\textwidth]{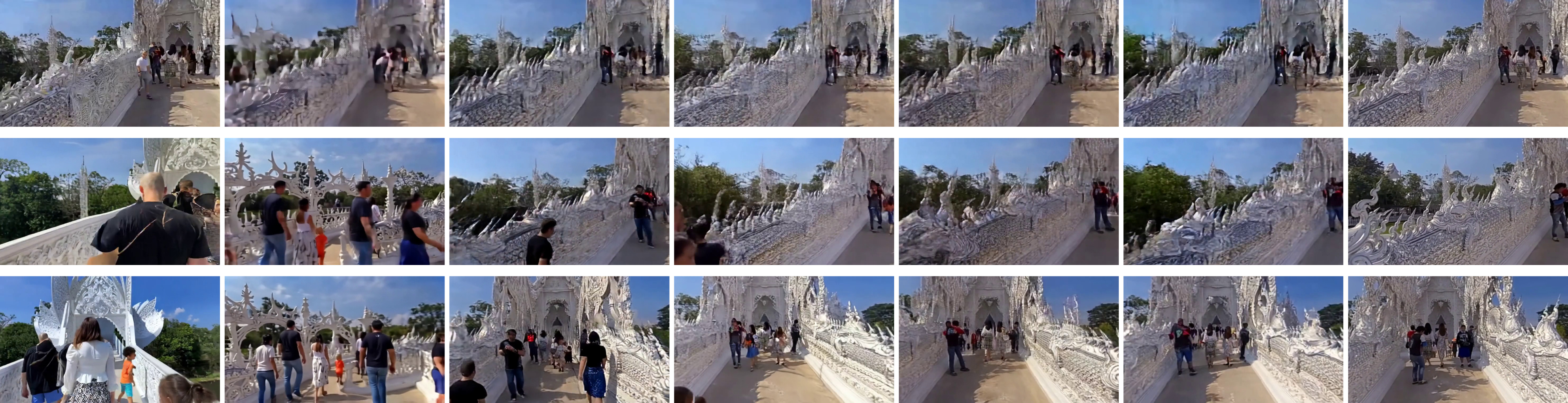}\par
\vspace{0pt}
{
\fontsize{7}{8}\selectfont
\setlength{\tabcolsep}{0pt}
\begin{tabular}{@{}*{7}{>{\centering\arraybackslash}p{\dimexpr\textwidth/7\relax}}@{}}
LiveWorld & SANA-WM & CameraCtrl & UCPE & CameraCtrl+OmniPersist & UCPE+OmniPersist & Ground Truth \\
\end{tabular}
}
\vspace{-3mm}
\caption{Prompt overview: \textit{A video shows a group of people walking along a white bridge with intricate carvings under a clear blue sky.}}
\label{fig:supp-revisits-VC-1}
\end{subfigure}\par
\vspace{3mm}
\begin{subfigure}[b]{\textwidth}
\centering
\includegraphics[width=\textwidth]{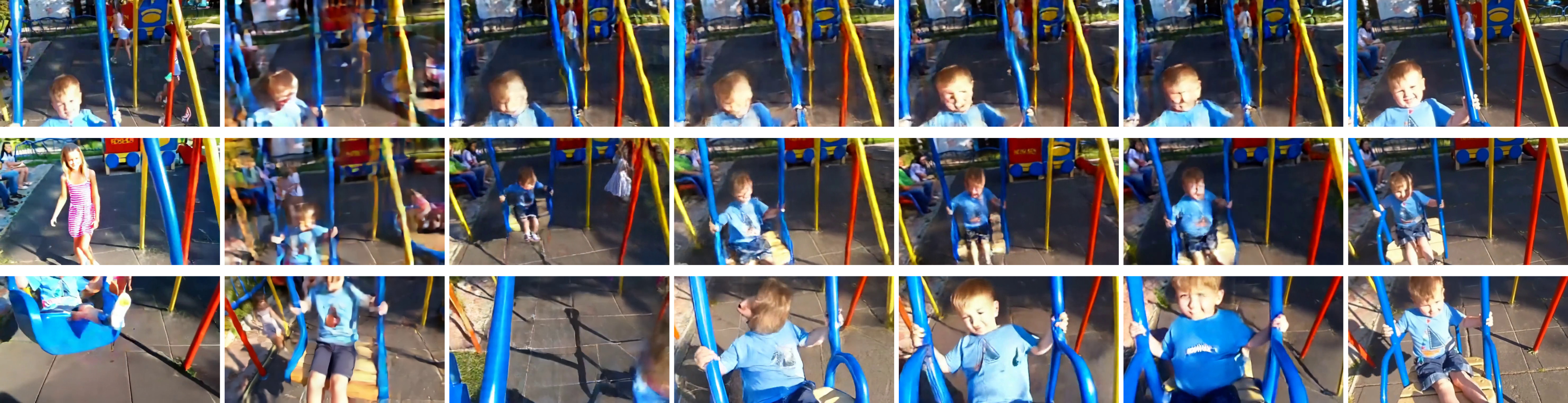}\par
\vspace{0pt}
{
\fontsize{7}{8}\selectfont
\setlength{\tabcolsep}{0pt}
\begin{tabular}{@{}*{7}{>{\centering\arraybackslash}p{\dimexpr\textwidth/7\relax}}@{}}
LiveWorld & SANA-WM & CameraCtrl & UCPE & CameraCtrl+OmniPersist & UCPE+OmniPersist & Ground Truth \\
\end{tabular}
}
\vspace{-3mm}
\caption{Prompt overview: \textit{A video shows a young boy in a blue shirt sitting on a swing, moving back and forth while holding the blue frame.}}
\label{fig:supp-revisits-VC-2}
\end{subfigure}\par
\vspace{1mm}
\caption{\textbf{Generation results on video continuation across camera revisits.} From top to bottom in each example, we show a frame before the camera turns away and two revisit frames after it returns to illustrate identity preservation and continued motion. Frames in each row share the same frame index across all methods and ground truth.}
\label{fig:supp-revisits-VC}
\end{figure*}

\noindent\textbf{Perspective Camera Revisits.}
Due to space limitations, we include only Figure~3 in the main paper for qualitative comparison and present additional results here. Because LiveWorld~\cite{duan2026liveworld}, LingBot-World 2.0~\cite{gao2026infinite} (for I2V), and SANA-WM~\cite{zhu2026sana} (for video continuation) perform relatively well among the released models, we select these models for comparison with the CameraCtrl~\cite{he2025cameractrl} and UCPE~\cite{zhang2026ucpe} baselines, as well as their corresponding CameraCtrl+OmniPersist ($s=4$) and UCPE+OmniPersist ($s=4$) variants.
Figures~\ref{fig:supp-revisits-I2V} and~\ref{fig:supp-revisits-VC} compare a pre-departure observation with early and later post-return observations at identical frame indices across these methods. The examples illustrate a consistent background, subject appearance, and continued motion after re-entry.  Specifically, Figure~\ref{fig:supp-revisits-I2V} presents I2V comparisons for two skaters and two vehicles, while Figure~\ref{fig:supp-revisits-VC} provides video continuation comparisons for people walking on a bridge and a boy on a swing.

\begin{figure*}[t!]
\centering
\captionsetup[subfigure]{font=promptoverview,skip=2pt}
\begin{subfigure}[b]{\textwidth}
\centering
\includegraphics[width=\textwidth]{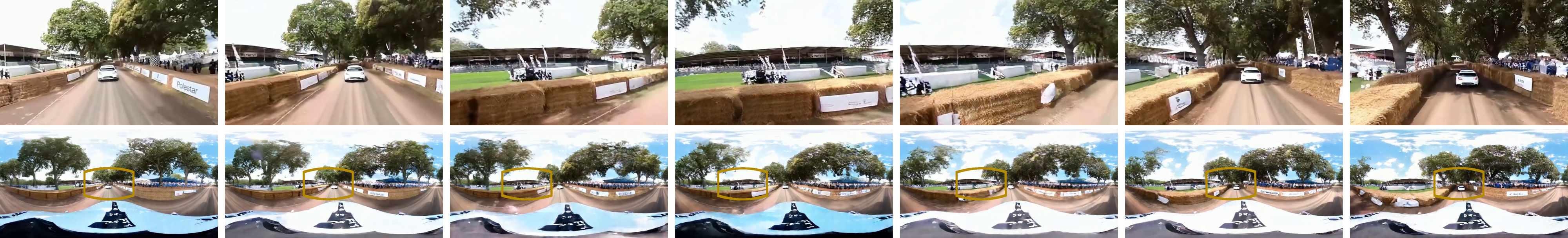}\par
\caption{Prompt overview: \textit{A video shows a white car driving forward on a dirt track lined with hay bales.}}
\label{fig:supp-panoramas-1}
\end{subfigure}\par
\vspace{1pt}
\begin{subfigure}[b]{\textwidth}
\centering
\includegraphics[width=\textwidth]{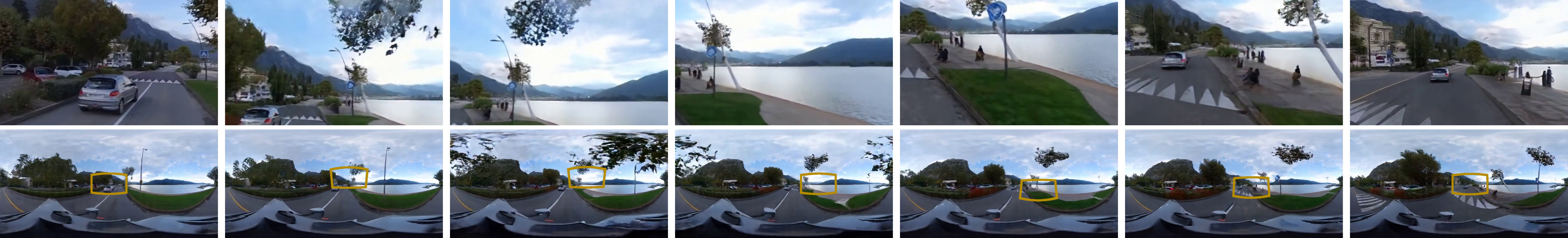}\par
\caption{Prompt overview: \textit{A video shows a silver hatchback car driving away from the camera on a paved road.}}
\label{fig:supp-panoramas-2}
\end{subfigure}\par
\vspace{1pt}
\begin{subfigure}[b]{\textwidth}
\centering
\includegraphics[width=\textwidth]{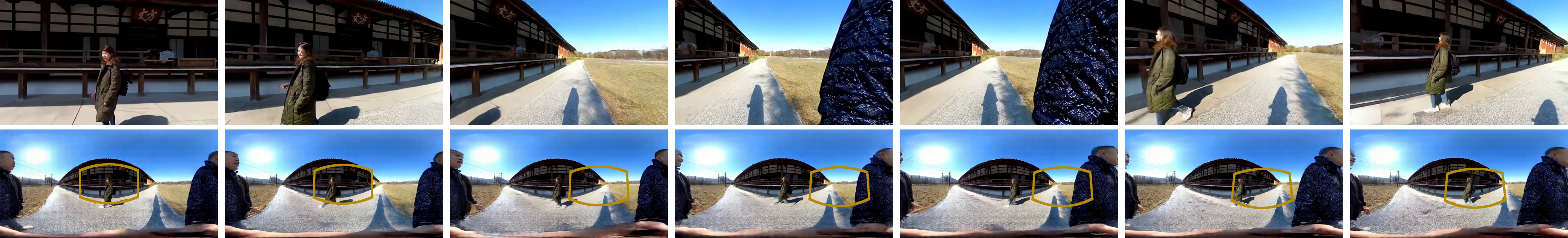}\par
\caption{Prompt overview: \textit{A video shows a woman with shoulder-length brown hair walking along a paved path next to a long, traditional wooden building with a dark roof.}}
\label{fig:supp-panoramas-3}
\end{subfigure}\par
\vspace{1pt}
\begin{subfigure}[b]{\textwidth}
\centering
\includegraphics[width=\textwidth]{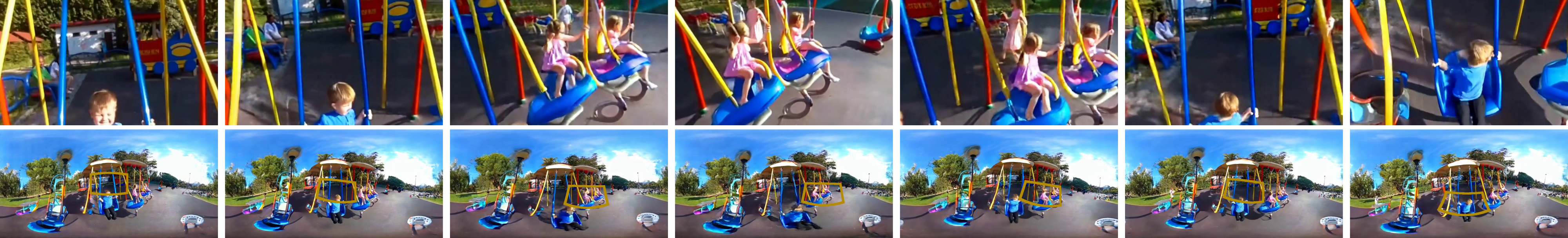}\par
\caption{Prompt overview: \textit{A video shows a young boy in a blue shirt sitting on a swing, moving back and forth while holding the blue frame.}}
\label{fig:supp-panoramas-4}
\end{subfigure}\par
\vspace{1pt}
\begin{subfigure}[b]{\textwidth}
\centering
\includegraphics[width=\textwidth]{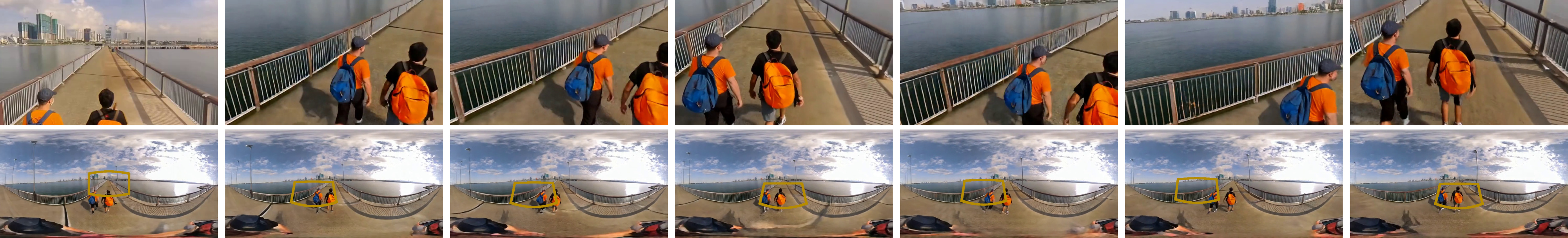}\par
\caption{Prompt overview: \textit{A video shows two men walking away from the viewer along a wide concrete pier.}}
\label{fig:supp-panoramas-5}
\end{subfigure}\par
\vspace{1pt}
\begin{subfigure}[b]{\textwidth}
\centering
\includegraphics[width=\textwidth]{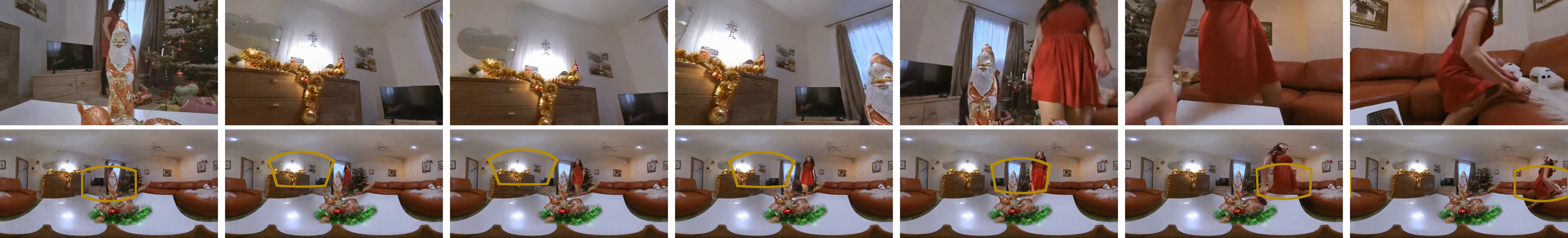}\par
\caption{Prompt overview: \textit{A video shows a woman in a reddish-orange dress standing in a living room decorated for Christmas. The woman finally sits on the sofa.}}
\label{fig:supp-panoramas-6}
\end{subfigure}\par
\caption{\textbf{Joint perspective generation and panoramic evolution under CameraCtrl+OmniPersist ($s=4$) on the I2V task.} Each subfigure shows generated perspective frames at $t=0,16,28,40,52,64,80$ in the top row and the corresponding panoramic frames in the bottom row. With panoramic sampling interval $s=4$, the panoramic-video frame indices are $r=0,4,7,10,13,16,20$, respectively, satisfying $t=sr$. Each perspective frame corresponds to the gold-boxed region in its synchronized panorama.}
\label{fig:supp-panoramas}
\end{figure*}

\noindent\textbf{Panoramic Evolution.}
Figure~\ref{fig:supp-panoramas} shows the generation results of CameraCtrl+OmniPersist at $s=4$ on the I2V task. We observe that the panoramic stream continues to evolve even when moving subjects leave the perspective field of view. Across the six examples, the panoramic states retain these subjects and depict changes in their positions during out-of-view intervals, providing evolving scene context for subsequent perspective revisits.


\section*{LLM Usage Statement}

During the preparation of this manuscript, Gemini 3.8 Flash was used for language polishing, and GPT-6 was employed for code debugging. 
The authors have reviewed all content and retain full responsibility for all methodological claims, implementation descriptions, experimental results, references, and the final text.

\clearpage
\clearpage

{\small\bibliographystyle{ieeenat_fullname}\bibliography{references}}

\end{document}